\documentclass[10pt,twocolumn,letterpaper]{article}
\usepackage{cvpr}              
\usepackage{graphicx}
\usepackage{amsmath}
\usepackage{amssymb}
\usepackage{booktabs}
\usepackage{multirow}
\usepackage{xcolor}
\usepackage{colortbl}
\usepackage{adjustbox}
\usepackage{float}

\usepackage[pagebackref,breaklinks,colorlinks]{hyperref}

\usepackage[capitalize]{cleveref}
\crefname{section}{Sec.}{Secs.}
\Crefname{section}{Section}{Sections}
\Crefname{table}{Table}{Tables}
\crefname{table}{Tab.}{Tabs.}

\def\cvprPaperID{10452} 
\def\confName{CVPR}
\def\confYear{2024}

\begin{document}

\title{PBWR: Parametric Building Wireframe Reconstruction from Aerial LiDAR Point Clouds}

\author{
Shangfeng Huang\textsuperscript{1}, Ruisheng Wang\textsuperscript{1,*}, Bo Guo\textsuperscript{2}, Hongxin Yang\textsuperscript{1} \\
\textsuperscript{1}University of Calgary, \textsuperscript{2}Guangdong University of Technology \\
\{\tt\small shangfeng.huang, ruiswang, hongxin.yang\}@ucalgary.ca, {\tt\small guobo.lidar@gmail.com}
}
\maketitle

\begin{abstract}
In this paper, we present an end-to-end 3D building wireframe reconstruction method to regress edges directly  from aerial LiDAR point clouds.
Our method, named Parametric Building Wireframe Reconstruction (PBWR), takes aerial LiDAR point clouds and initial edge entities as input, and fully uses self-attention mechanism of transformers to regress edge parameters without any intermediate steps such as corner prediction. We propose an edge non-maximum suppression (E-NMS) module based on edge similarityto remove redundant edges. Additionally, a dedicated edge loss function is utilized to guide the PBWR in regressing edges parameters, where simple use of edge distance loss isn't suitable.  In our experiments, we demonstrate state-of-the-art results on the Building3D dataset, achieving an improvement of approximately 36\% in entry-level dataset edge accuracy and around 42\% improvement in the Tallinn dataset.
\end{abstract}

\section{Introduction}
\label{sec:intro}

3D Building wireframe reconstruction is an important mid-level visual process\cite{man1982computational}. 3D wireframe models, being lightweight in nature, offer a comprehensive representation of shape and structural information of 3D objects. Further, they can be easily and efficiently converted into 3D mesh models including CAD models that play a crucial role in metaverse, smart cities, and virtual reality applications etc. 
Despite its practical and scientific importance, 3D building wireframe reconstruction still remains an unsolved problem in computer vision. 
\begin{figure}[htbp]
    \centering
    \includegraphics[width=0.95\linewidth]{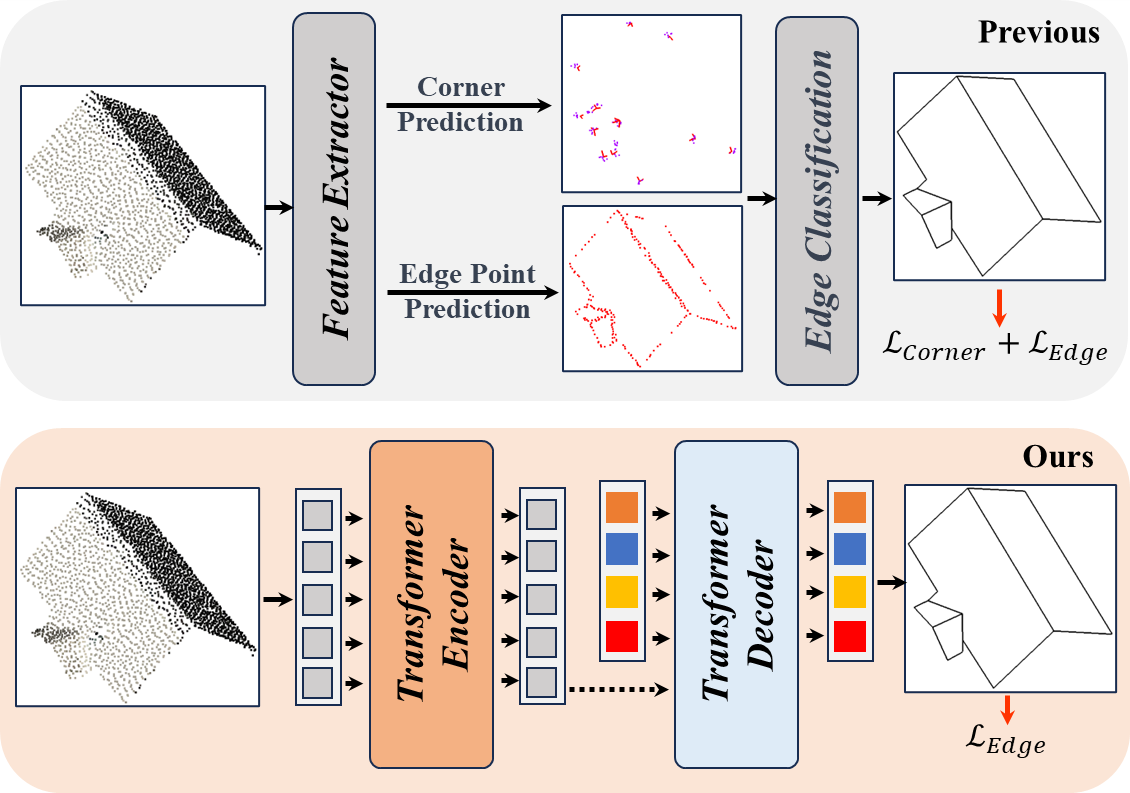}
    \caption{\textbf{Pipeline Comparison}. \textbf{Top:} The most popular wireframe reconstruction method based on point prediction and edge classification modules, are optimized by $ \mathcal{L}_{Corner}$ and $ \mathcal{L}_{Edge}$, respectively.\textbf{ Bottom:} Our proposed PBWR  avoids the intermediate point classification process.}
    \label{fig:Pipeline Comparison}
\end{figure}

In general, point clouds are considered as one of the simplest and most flexible ways to represent 3D objects. It possesses innate ability to represent structure of objects with remarkable fidelity, capturing intricate details with precision. Wireframe models, on the other hand, are also specifically designed to represent 3D objects. Unlike meshes, wireframes are more simplified representation of 3D objects and can be readily converted back and forth into CAD models. Therefore, ABC \cite{Koch_2019_CVPR} and Building3D \cite{Wang_2023_ICCV} datasets recommend the use of wireframe models for point cloud reconstruction. Existing state-of-the-art deep learning methods \cite{liu2021pc2wf, li2022point2roof, jiang2023extracting,zhu2023nerve} demonstrate impressive performance in wirefame reconstruction, but they are limited by using dense synthetic point clouds. Additionally, they involve heuristic-guided intermediate processing modules such as corner prediction and edge classification \cite{Wang_2023_ICCV, liu2021pc2wf,li2022point2roof}, which lead to error accumulation and limit performance improvement and further development.

In this paper, we propose a novel approach named\textbf{ }Parametric Building Wireframe Reconstruction from aerial LiDAR point clouds (PBWR).
It is the first time to directly regress edges from point clouds using transformers\cite{vaswani2017attention} and eliminate intermediate steps such as corner prediction and edge classification.
Existing methods \cite{liu2021pc2wf, li2022point2roof, Wang_2023_ICCV}  predict $N$ candidate corners that can generate $C_N^2$ edge proposals, where approximately 85\% of the edges are considered useless. The PBWR is proposed to directly regress desired edges to mitigate generation of a large number of extraneous edges, as described in the ablation study.



On the other hand, these methods \cite{liu2021pc2wf, li2022point2roof, Wang_2023_ICCV} simply rely on endpoint distances between edges for bipartite matching, often resulting in significant matching errors. 
We propose a novel edge similarity method that incorporates Hausdorff distance \cite{hausdorff1918dimension}, edge length and cosine similarity to enhance the bipartite matching, which significantly improve the performance of edge regression. 
Furthermore, an edge non-maximum suppression (E-NMS) is designed to eliminate redundant positive edges, and a dedicated edge loss function is proposed to guide the strategy of directly regressing edges.

\textbf{Contributions} This paper contributes three main advancements to the field of wireframe reconstruction.
\begin{itemize}
    \item We propose a novel method called PBWR, which directly regresses desired edges without any intermediate heuristic-guided process such as corner prediction and edge classification, and yields a substantial number of positive candidate edges.
\item It is the first time to propose edge similarity (\cref{eq_edge_similarity}) and dedicated edge loss functions in point clouds, effectively guiding reconstruction of wireframe models. An edge non-maximum suppression method based on the edge similarity is proposed for the first time, applied to the wireframe reconstruction task.
\item Extensive experiments demonstrate that the PBWR brings a significant improvement in wireframe reconstruction compare to these methods with intermediate heuristic steps. For example, the edge recall rate has a notable improvement in Entry-level (from 46\% to 82\%) and Tallinn city (from 23\% to 65\%) data.
\end{itemize}

\section{Related Work}
\label{sec:related}

\subsection{Building Reconstruction}  Many optimization-based algorithms \cite{chen2017topologically, nan2017polyfit, he2021manhattan, fang2020connect, lin2013semantic, holzmann2018semantically} have been proposed for building reconstruction from point clouds. The primitive-based building reconstruction 
\cite{he2021manhattan,wang2020robust,chen2017topologically,nan2017polyfit, zhou20102, fang2020connect} is a popular method for generating polygonal meshes. Another widely adopted  building reconstruction methods \cite{he2021manhattan, li2016manhattan, verma20063d, lafarge2011building} are based on strong Manhattan World  assumption \cite{coughlan2000manhattan} that enforces planes to follow only three orthogonal directions to generate dense meshes. City3D \cite{huang2022city3d} combines initial meshes generated by PolyFit \cite{nan2017polyfit} with vertical walls from height maps to reconstruct buildings from aerial LiDAR point clouds. 
Chen et al. \cite{chen2017topologically} identified rooftop boundaries from aerial LiDAR point clouds and clustered them in terms of topological consistency 
,and directly generated vertical walls connecting roofs. Similarly, 2.5D dual contouring \cite{zhou20102}, which extends classic dual contouring into a 2.5D method \cite{ju2002dual}, employs an adaptive grid as supporting data structure   to infer roofs in each grid node. 
These traditional methods demonstrate effectiveness in practice, but they accumulate noticeable errors during the inference process. In addition, model's sensitivity to parameters can lead to unstable results. 

\begin{figure*}
    \centering
    \includegraphics[width=1\linewidth]{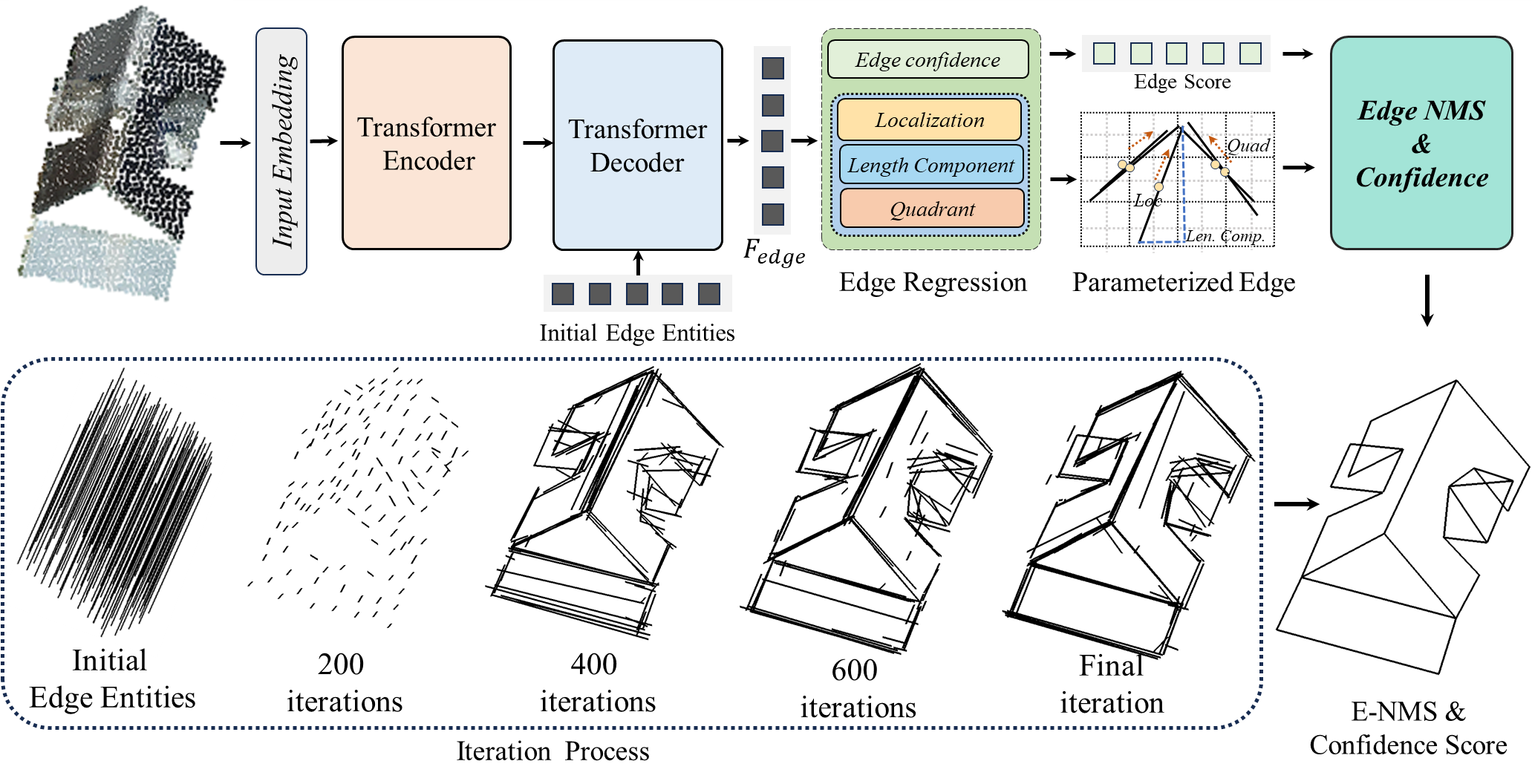}
    \caption{\textbf{ PBWR Overview}. \textbf{Top}: the illustration of PBWR pipeline. It takes a roof point cloud as input and uses Input Embedding and Transformer Encoder modules to generate high-dimensional features for each point. Subsequently, point features and initial edge entities as query embeddings, are fed into the Transformer Decoder and Edge Regression modules to obtain edge regression results, then optimized by the E-NMS to generate the wireframe model. \textbf{Bottom}: the reconstruction results are presented at different iterations during training.}
    \label{fig_pipeline}
\end{figure*}

\subsection{Wireframe Reconstruction}
As deep learning in point cloud processing continues to flourish\cite{qi2017pointnet++, zhao2021point, yu2022point}, there is a growing exploration of its applications in wireframe reconstruction. Initially, some studies  \cite{yu2018ec, zhang2020large, himeur2021pcednet} have attempted to treat wireframe reconstruction as an edge point classification task, where these edge points, also known as contour points, can also represent geometric structure of objects. 
Specifically, EC-Net \cite{yu2018ec} identifies object contour by reconstructing distance distribution from each point to the edges.
However, this representation isn't practical. To construct solid wireframe models, recent research \cite{liu2021pc2wf, li2022point2roof, jiang2023extracting} first predict positions of corners then regress edges from a large number of edge proposal, as shown in \cref{fig:Pipeline Comparison}.  
Specifically, PC2WF \cite{liu2021pc2wf} initially refines corner positions from generated corner candidate patches.
NerVE \cite{zhu2023nerve} classifies contour voxels from voxelized point clouds to infer edge parameters. However, obtaining accurate corner or edge coordinates from complex scenes can frequently be error-pone, limiting performance of  subsequent edge regression. Therefore, the proposed PBWR uses a powerful transformer to directly regress edges without  the need to predict corners or edge proposals. 


\section{The Method}
\label{sec:method}
\subsection{Motivation}
\label{sec:motivation}
When consider direct regression of an edge in 3D space, a straightforward method is to predict two endpoints of an edge. However, existing methods \cite{qi2019deep, misra2021end} have demonstrated that directly regressing precise positions of points in 3D space is extremely challenging. 
Another alternative method is to determine an edge by regressing its parameters. In this work, parameterized edges are formulated as follows: 

\begin{equation}\label{eq_parameter_edge}
\begin{aligned}
  \begin{bmatrix}V_i\\V_j\end{bmatrix} &= \begin{bmatrix}1 & 1/2 \\ 1 & -1/2
\end{bmatrix} \begin{bmatrix}P_m \\ \vec{v}\end{bmatrix}
= \begin{bmatrix}1 & 1 \\ 1 & -1
\end{bmatrix} \begin{bmatrix}P_m \\ \vec{u}\odot v' \end{bmatrix}\\
  &=\begin{bmatrix}1 & 1/2 \\1 & -1/2\end{bmatrix}
\left ( \begin{bmatrix}
 I_{3*1}\\\vec{u}_{3*1} 
\end{bmatrix} \odot \begin{bmatrix}x_m&y_m&z_m \\\left |  v_i \right |  & \left | v_j \right |  & \left | v_z \right |  \end{bmatrix} \right ) 
\end{aligned}
\end{equation}
where $V_i$ and $V_j$ represent two endpoints of an edge,  $P_m$ denotes midpoint of the edge,  $\vec{v}$ denotes the directional vector equal to $V_i - V_j$. 
Additionally, $\vec{u}$ represents the symbolic matrix,  $I$ denotes an identity matrix, and  $v' = [\left |  v_i \right |, \left | v_j \right |, \left | v_z \right |]$ denotes projected length of the edge along XYZ axes.
An essential parameter of the parameterized edge is a known point on the edge.
Compared to other points on an edge, regressing the midpoint $P_m$ is a preferable choice to locate position of the edge. 
Other two crucial parameters are directional vector and length of the edge. The directional vector $\vec{v}$ can be decomposed into directional information $\vec{u}$ and projected lengths $v'$ of the edge. Directional information $\vec{u}$, also known as the symbolic matrix,  essentially indicates the quadrant towards which the edge is pointing. 
As edges are directionless, $\vec{u}$ can be represented by four rather than eight quadrants. 
PBWR regresses edges from roof point clouds by predicting these seven parameters. We also propose PBWR-Corner that directly regress endpoints to determine the edge as a comparative approach.   

\subsection{Overall Network Architecture}
In this work, PBWR takes roof point clouds as input and predicts parameterized edges, then fed into the E-NMS \& Confidence module to reconstruct a wireframe model as shown in \cref{fig_pipeline}. For a given point cloud with $N$ points $P \in \mathbb{R}^{N\times7}$ encompassing coordinates, RGB information and reflection, it is initially directed to the input embedding module, where a multi-layer perceptron (MLP)  is employed to aggregate local structural information of points. The resulting features $F_{embed} \in \mathbb{R}^{N\times C_{embed}}$ are input to the Transformer Encoder module to obtain $F_{en} \in \mathbb{R}^{N\times C_{en}}$ features. Subsequently, initial edge entities for edge queries and $F_{en}$ are passed into the Transformer Decoder module for learning edge representation. Finally, the resulting edge features $F_{edge}$ are parsed by the Edge Regression module and  redundant edges are removed by E-NMS. The results of edges at different iteration stages are also presented in the \cref{fig_pipeline} (bottom).

\subsection{Transformer Encoder and Decoder}
In this work, the standard Transformer with self-attention mechanism \cite{vaswani2017attention} is employed. 
The Transformer Encoder and Decoder modules as shown in \cref{fig_pipeline} haven't any specific modifications to adapt for 3D data. Even the sampling strategies, widely employed in 3D Transformer mechanisms \cite{zhao2021point, lai2022stratified, guo2021pct} to reduce computational overhead and enhance performance, has not been used in our work. 
In PBWR, the Transformer Decoder is designed to learn distinguishable edge features, along with its input , which comprises encoder features $F_{en}$ and initial edges entities, where initialized entities are considered as object queries.  Generally, object queries are often kept consistent with the input encoder module, such as point coordinate queries \cite{lai2022stratified, zhao2021point} and point patch queries \cite{yu2022point, pang2022masked}. 
Similarly, we choose $M$ query points using the Farthest Point Sampling strategy as object queries. The initial edge entities entail object queries with an initial direction and length. The M query points frequently doesn't align with the midpoints of the edges.
Inspired by recent 3D object detection works which predict residual offsets from original candidate points to interesting points\cite{qi2019deep, misra2021end}, PBWR predicts the distance from query points to the nearest midpoint of the edges. Specifically,  $M$ query points are input into positional embedding \cite{vaswani2017attention} to obtain query embeddings.
The resulting query embeddings $F_{query} \in \mathbb{R}^{M \times C_{query}}$ and point features $F_{en}$ are fed into the Transformer Decoder to generate distinguishable edge features $F_{edge} \in \mathbb{R}^{M \times C_{edge}}$. 
In the Edge Regression module, $F_{edge}$ are parsed to obtain seven parameters for edges, including residual offsets. 

\subsection{Edge Regression Module}
The Edge Regression module is designed to parse the features $F_{edge}$ obtained from the Transformer Encoder and Decoder structures to generate parameterized edges. It employs three dedicated MLPs to regress distinct parameters of the parameterized edges, specifically $P_m$, $v'$ and $\vec{u}$, as described in \cref{eq_parameter_edge}.  The localization MLP predicts residual offsets from candidate points to midpoints of edges instead of predicting midpoints directly. The orientation of the edge is collectively determined by the absolute values $v'$ of the components of the edge along the XYZ axes and $\vec{u}$ serving as the symbolic matrix. The symbolic matrix can be regards as a classification problem, specifically determining which quadrant the vector points towards, as described in \cref{sec:motivation}. Therefore, a component MLP is employed to predict three component values, while a quadrant MLP is utilized to predict probability of the predicted edge vector pointing from various quadrants, simply named as quadrant classes. Additionally, an additional MLP is employed to obtain confidence scores for the edges. 

\subsection{Bipartite Edge Matching and E-NMS}
\begin{figure}[htbp]
    \centering
    \begin{subfigure}[t]{0.46\linewidth}
           \centering
           \includegraphics[width=1\linewidth]{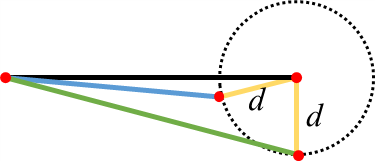}
            \caption{}
            \label{fig_ENMS_a}
    \end{subfigure}
    \hfill
    \begin{subfigure}[t]{0.46\linewidth}
            \centering
            \includegraphics[width=1\linewidth]{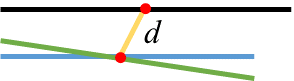}
            \caption{}
            \label{fig_ENMS_b}
    \end{subfigure}
    \caption{(a) edge distance $d$ quantified by corner distance, and (b) edge distance $d$ quantified by midpoint distance.}
    \label{fig_ENMS}
\end{figure}
Effectively bipartite matching between prediction and ground truth is crucial in deep learning. Current methods \cite{liu2021pc2wf, Wang_2023_ICCV, li2022point2roof} match prediction with ground truth by minimizing distance between predicted corners and ground truth corners, which naturally extends to matching edges. However, it's not applicable to the proposed PBWR, as it circumvents the intermediate step of predicting corners. Hence, a specialized bipartite edge matching strategy is proposed to address wireframe reconstruction problem. It's evident that distance, direction, and length can be employed to quantify similarity between edges. \textbf{Distance:} The \cref{fig_ENMS_a} illustrates that use of corner distance  results in same distances from blue and green to black edges. The midpoint distance also yields the same distance in the example shown in \cref{fig_ENMS_b}. However, the blue edge is a more suitable match when considering the black edge. Therefore, the Hausdorff distance defining the distance between two point sets is employed to quantify edge distance for optimal bipartite edge matching:
\begin{equation}\label{eq_Hausdorff}
\begin{aligned}
  H_d\left(e_i, e_j\right) &= \max\left(h_d\left( e_i,e_j\right), h_d\left(e_j, e_i\right) \right) \\
  h_d\left( e_i,e_j\right) &= \max_{p_{e_i}\in e_i} \left \{\min_{p_{e_j} \in e_j} \left \| p_{e_i} - p_{e_j} \right \| \right \} \\
  h_d\left(e_j, e_i\right) &= \max_{p_{e_j}\in e_j} \left \{\min_{p_{e_i} \in e_i} \left \| p_{e_j} - p_{e_i} \right \| \right \}
\end{aligned}
\end{equation}
where $p_{e_i}$ and $p_{e_j}$ represent points sampled uniformly from edges $e_i$ and $e_j$, respectively. $  H_d\left(e_i, e_j\right)$ denotes the quantified distance between edges $e_i$ and $e_j$. \textbf{Direction:} Cosine similarity is utilized to evaluate the similarity in edge. \textbf{Length: } Length similarity is quantified based on the ratio of lengths between edges. They can be formulated as follows:
\begin{equation}
\begin{aligned}
  Dir_{sim} \left(e_i, e_j \right) &= 1 - \frac{\left |e_i \cdot e_j\right | }{\left\| e_i\right\| \times \left\| e_j \right\|} \\
  Len_{sim} \left(e_i, e_j \right) &= 1 - \frac{\min(\left\| e_i \right\|, \left\| e_j \right\|)}{\max(\left\| e_i \right\|, \left\| e_j \right\|)}
\end{aligned}
\end{equation}
Hence, the \textbf{edge similarity} can be represented as follows:
\begin{equation}\label{eq_edge_similarity}
\begin{aligned}
  Edge_{sim}\left(e_i, e_j \right) &= \alpha H_d\left(e_i, e_j\right) + \beta  Dir_{sim} \left(e_i, e_j \right)\\  &+ \gamma Len_{sim}\left(e_i, e_j \right)
\end{aligned}
\end{equation}
where $\alpha$, $\beta$, $\gamma$ denote the balancing coefficients. This indicates that when the $Edge_{sim}\left(e_i, e_j \right)$ value is close to 0,  edges $e_i$ and $e_j$ are similar. The Hausdorff distance also implicitly consider edge length and directional information, as evidenced by the \cref{eq_Hausdorff} and the selection of the better-matching blue edge in \cref{fig_ENMS} based on the Hausdorff distance. However, a reasonable combination of all three factors contributes to improve performance, as demonstrated in the ablation study. The E-NMS algorithm takes both edge similarity and predicted confidence scores  as inputs to remove redundant edges. Details of the E-NMS pseudocode can be found in the \textit{supplementary materials}.

\subsection{Loss Function}
$Edge_{sim}\left(P_{edge}, G_{edge} \right)$ can be utilized to calculate the edge similarity between predicted edge set $P_{edge}$ and ground truth edge set $G_{edge}$. The resulting edge similarity is fed into the Hungarian algorithm to perform bipartite edge matching between $N_{pos}$ positive predictions and the ground truth edge set.

\noindent \textbf{Midpoint and Component length loss}:  The $\ell_1$ distance loss is employed for both midpoint positions and predicted lengths of components $v'$ in the XYZ axis.  The midpoint loss $ \mathcal{L}_{mid}$ is formulated as:
\begin{equation}
\begin{aligned}
  \mathcal{L}_{mid} &= \frac{1}{N_{pos}}\sum_{i,j \in N_{pos}} \left(p_{mid}^i - g_{mid}^j\right) 
\end{aligned}
\end{equation}
where $p$ and $g$ represent elements within the sets of edges, $P_{edge}$ and $G_{edge}$, respectively. The component length loss $ \mathcal{L}_{comp}$ is similar to the midpoint loss.

\noindent \textbf{Confidence and Quadrant classification loss}: The $\ell_{CE}$ cross-entropy loss is used to optimize  predicted confidence scores and quadrant classes as follows:
\begin{equation}\label{equ_conf}
\begin{aligned}
  \mathcal{L}_{con} &= \ell_{CE} \left(p_{con}, g_{con}\right) \\
  g_{con} &= \begin{cases}
  1-Edge_{sim} &\text{ if } Edge_{sim} <1 \\
  0 &\text{ otherwise }
\end{cases} \\
\mathcal{L}_{quad} &= \frac{1}{N_{pos}} \, \underset{i,j \in N_{pos}}{\ell_{CE}} \left(p_{quad}^i, g_{quad}^j\right) \\
\end{aligned}
\end{equation}
When simply setting the confidence scores of positive predictions to 1 and negative predictions to 0, the extreme class imbalance leads to the network failing to converge, regardless of whether cross-entropy or Focal loss is used. Therefore, the confidence scores of the ground truth are redefined based on edge similarity $Edge_{sim}$. 

\noindent\textbf{Edge similarity loss $\mathcal{L}_{sim}$} is directly computed as average edge similarity scores for $N_{pos}$ positive predictions. Thus, the final loss $\mathcal{L}$ is formulated as: 
\begin{equation}
\begin{aligned}
  \mathcal{L} &= \lambda_{mid}\mathcal{L}_{mid}+\lambda_{comp}\mathcal{L}_{comp} + \lambda_{con}\mathcal{L}_{con}\\ &+ \lambda_{quad}\mathcal{L}_{quad} + \lambda_{sim}\mathcal{L}_{sim}
\end{aligned}
\end{equation}
where $\lambda_{*}$ is employed as a coefficient to balance various loss terms.

\section{Experiments}
\label{sec:expe}

\subsection{Dataset and Evaluation Metric}
\label{sec:metric}
Building3D \cite{Wang_2023_ICCV} dataset is employed to evaluate our model.  Specifically, the Entry-level dataset of Building3D consists of 5,698 training point clouds and 583 testing point clouds, while the Tallinn city dataset   includes 32,618 training point clouds and 3,472 testing point clouds. All the benchmark samples are extracted from sparse aerial point clouds. During training stages, data augmentation is implemented to enhance the network with desirable robustness and invariance. Specifically, We add augment of horizontal flips along the YZ or XZ plane, and random rotations along the Z-axis (-5\textdegree $\sim$ 5\textdegree), to increase the input shape diversity.
\begin{table*}
    \centering
    \begin{tabular}{c|c|c|ccc|ccc}
    \toprule
          \multicolumn{2}{c|}{\multirow{2}{*}{Method}}&  Distance (m)&  \multicolumn{6}{c}{Accuracy}\\
  \multicolumn{2}{c|}{}& ACO & CP& CR& $\text{CF}_1$& EP& ER&$\text{EF}_1$\\
    \midrule
         \multirow{8}{*}{Building3D}&PointNet* \cite{qi2017pointnet}&  0.36&  0.71&  0.50&  0.59&  0.81&  0.26&0.39\\
          &PointNet++* \cite{qi2017pointnet++}&  0.34&  0.79&  0.52&  0.63&  0.84&  0.33& 0.47\\
          &RandLA-Net* \cite{hu2020randla}&  0.35&  0.70&  0.60&  0.65&  0.67&  0.16& 0.25\\
          &DGCNN* \cite{phan2018dgcnn}&  0.32&  0.73&  0.58&  0.65&  0.81&  0.30& 0.44\\
          &PAConv*  \cite{xu2021paconv}&  0.33&  0.75&  0.57&  0.65&  0.85&  0.31& 0.45\\
          &Stratified Transformer* \cite{lai2022stratified}&  0.38&  0.72&  0.51&  0.62&  0.75&  0.22& 0.34\\
          &Point2Roof \cite{li2022point2roof}&  0.30&  0.66&  0.48&  0.56&  0.71&  0.26& 0.38\\
 & Building3D-supervised \cite{Wang_2023_ICCV}& 0.26& 0.89& 0.66& 0.76& 0.91& 0.46&0.61\\
 \midrule
 \multirow{2}{*}{Tested by Us}
 & PC2WF \cite{liu2021pc2wf} & 0.45& 0.24& 0.13& 0.16& 0.02& 0.14&0.04\\
          &\cellcolor{blue!8}PBWR&\cellcolor{blue!8}  0.22& \cellcolor{blue!8} 0.97&\cellcolor{blue!8}  0.85&\cellcolor{blue!8}  0.91& \cellcolor{blue!8} 0.91&\cellcolor{blue!8}  0.82 &\cellcolor{blue!8} 0.86\\
          &\cellcolor{blue!8}PBWR-Tallinn & \cellcolor{blue!8} \textbf{0.18}& \cellcolor{blue!8} \textbf{0.99}& \cellcolor{blue!8} \textbf{0.87}&\cellcolor{blue!8}  \textbf{0.93}& \cellcolor{blue!8} \textbf{0.96}&\cellcolor{blue!8}  \textbf{0.84 }&\cellcolor{blue!8} \textbf{0.90}\\
    \bottomrule
    \end{tabular}
    \caption{Performance comparisons were conducted on the Entry-level data from the Building3D dataset. Their accuracy was provided by the Building3D dataset or tested by us. * indicates that this method serves as the feature extractor in the wireframe reconstruction network. PBWR-Tallinn represents training on Tallinn city data, while evaluating on Entry-level data.}
    \label{tab_entry_level}
\end{table*}
The same evaluation metric \cite{Wang_2023_ICCV} are employed to evaluate the proposed PBWR as shown in \cref{tab_entry_level}. Average Corner Offset (ACO) metric is used to evaluate average offsets between predicted and ground truth corners. Corner Precision (CP) and Edge Precision (EP) represent the precision of corner prediction and edge prediction, respectively. The Corner Precision (CP) and Edge Precision (ER) denote the recall of corners and edges, respectively. Corner $\text{F}_1$ score ($\text{CF}_1$) and Edge $\text{F}_1$ score ($\text{EF}_1$) represent   corresponding $\text{F}_1$ scores.

\begin{table*}
    \centering
    \begin{tabular}{c|c|c|ccc|ccc}
    \toprule
          \multicolumn{2}{c|}{\multirow{2}{*}{Method}}&  Distance (m)&  \multicolumn{6}{c}{Accuracy (\%)}\\
  \multicolumn{2}{c|}{}& ACO & CP& CR& $\text{CF}_1$& EP& ER&$\text{EF}_1$\\
    \midrule
         \multirow{5}{*}{Building3D}&PointMAE*\cite{qi2017pointnet}&  0.33&  0.75&  0.47&  0.58&  0.52&  0.12&0.20\\
          &PointM2AE* \cite{qi2017pointnet++}&  0.32&  0.79&  0.58&  0.67&  0.50&  0.07& 0.12\\
          &Point2Roof \cite{hu2020randla}&  0.39&  0.65&  0.30&  0.41&  0.66&  0.08& 0.14\\
          &Building3D-Linear self-supervised\cite{phan2018dgcnn}&  0.35&  0.70&  0.60&  0.65&  0.67&  0.16& 0.25\\
          &Building3D-supervised \cite{Wang_2023_ICCV}&  0.29&  0.90&  0.53&  0.66&  0.88&  0.23& 0.36\\
 \midrule
 \multirow{2}{*}{Tested by Us}& PC2WF \cite{liu2021pc2wf} & 0.52& 0.18& 0.67& 0.28& 0.02& 0.15&0.01\\
          &\cellcolor{blue!8}PBWR-Tallinn& \cellcolor{blue!8} \textbf{0.22}&\cellcolor{blue!8}  \textbf{0.96}&  \cellcolor{blue!8}\textbf{0.68}&\cellcolor{blue!8}  \textbf{0.80}&\cellcolor{blue!8}  \textbf{0.91}& \cellcolor{blue!8} \textbf{0.65}&\cellcolor{blue!8} \textbf{0.76}\\
    \bottomrule
    \end{tabular}
    \caption{Performance comparisons were conducted on the Tallinn city data from the Building3D dataset. Their accuracy was provided by the Building3D dataset or tested by us. * indicates that this method serves as the feature extractor in the wireframe reconstruction network.}
    \label{tab_Tallinn}
\end{table*}

\begin{figure*}
    \centering
    \begin{minipage}[t]{0.08\linewidth}
    \vspace{0pt}
        \includegraphics[width=\linewidth]{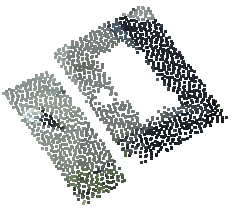}
        \includegraphics[width=\linewidth]{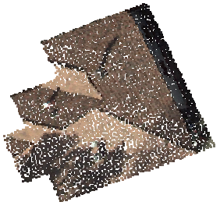}
        \includegraphics[width=\linewidth]{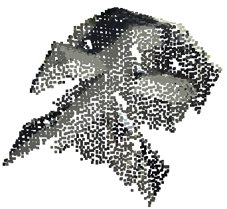}
        \includegraphics[width=\linewidth]{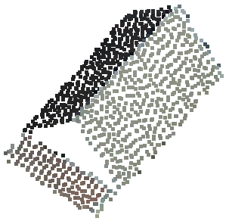}
        \includegraphics[width=\linewidth]{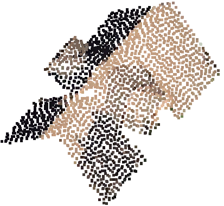}
        \includegraphics[width=\linewidth]{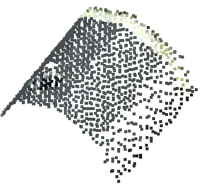}
        \includegraphics[width=\linewidth]{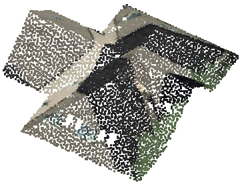}
        \includegraphics[width=\linewidth]{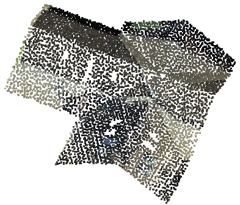}
        \includegraphics[width=\linewidth]{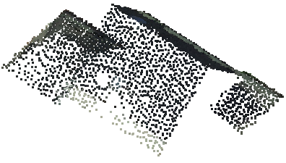}
        \subcaption*{\parbox{\linewidth}{\centering Point  Cloud}}
    \end{minipage}
    \hfill
    \begin{minipage}[t]{0.08\linewidth}
    \vspace{0pt}
        \includegraphics[width=\linewidth]{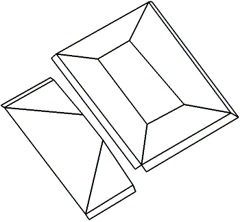}
        \includegraphics[width=\linewidth]{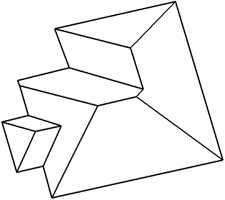}
        \includegraphics[width=\linewidth]{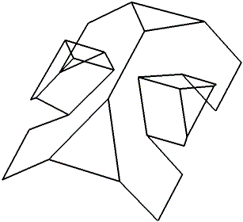}
        \includegraphics[width=0.9\linewidth]{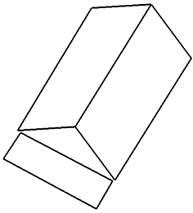}
        \includegraphics[width=0.9\linewidth]{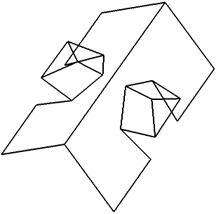}
        \includegraphics[width=0.8\linewidth]{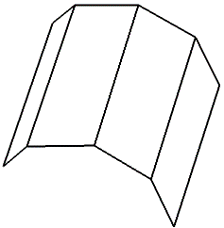}
        \includegraphics[width=0.85\linewidth]{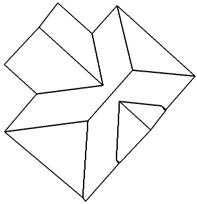}
        \includegraphics[width=0.8\linewidth]{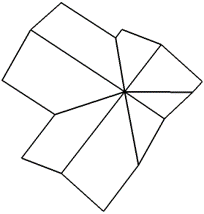}
        \includegraphics[width=\linewidth]{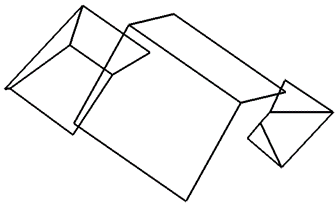}
        \subcaption*{\parbox{\linewidth}{\centering  Wireframe}}
    \end{minipage}
    \hfill
    \begin{minipage}[t]{0.08\linewidth}
    \vspace{0pt}
        \includegraphics[width=0.9\linewidth]{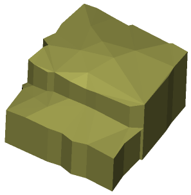}
        \includegraphics[width=\linewidth]{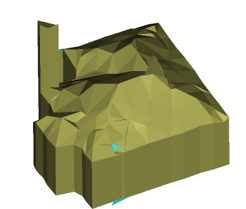}
        \includegraphics[width=0.95\linewidth]{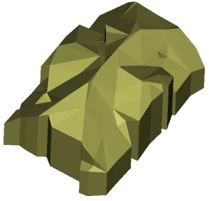}
        \includegraphics[width=0.9\linewidth]{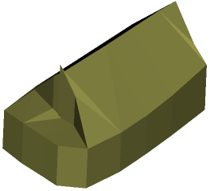}
        \includegraphics[width=0.9\linewidth]{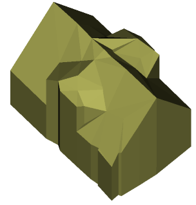}
        \includegraphics[width=0.85\linewidth]{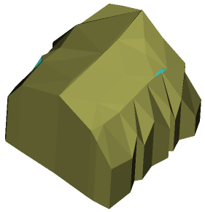}
        \includegraphics[width=0.9\linewidth]{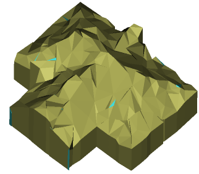}
        \includegraphics[width=0.9\linewidth]{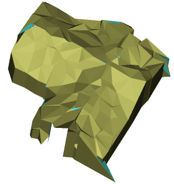}
        \includegraphics[width=0.9\linewidth]{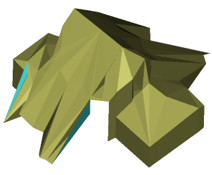}
        \subcaption*{\parbox{\linewidth}{\centering 2.5D Dual \newline \cite{zhou20102}}}
    \end{minipage}
    \hfill
    \begin{minipage}[t]{0.08\linewidth}
    \vspace{0pt}
        \includegraphics[width=0.9\linewidth]{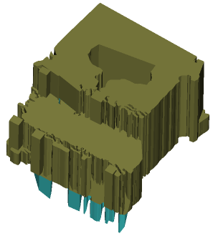}
        \includegraphics[width=0.78\linewidth]{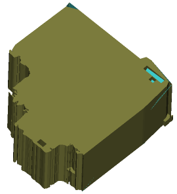}
        \includegraphics[width=0.95\linewidth]{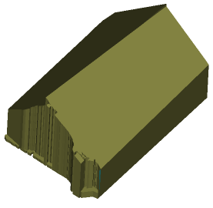}
        \includegraphics[width=0.85\linewidth]{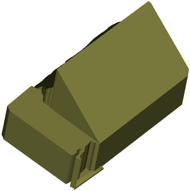}
        \includegraphics[width=\linewidth]{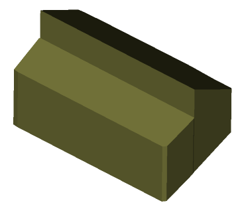}
        \includegraphics[width=0.85\linewidth]{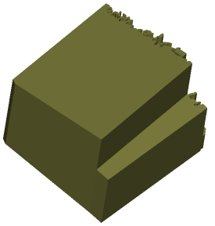}
        \includegraphics[width=0.85\linewidth]{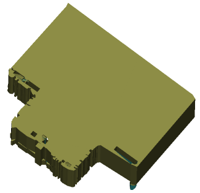}
        \includegraphics[width=0.9\linewidth]{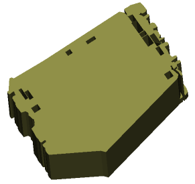}
        \includegraphics[width=0.9\linewidth]{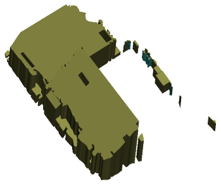}
        \subcaption*{\parbox{\linewidth}{\centering Topology \newline Aware \cite{chen2017topologically}}}
    \end{minipage}
    \hfill
    \begin{minipage}[t]{0.08\linewidth}
    \vspace{0pt}
        \includegraphics[width=0.9\linewidth]{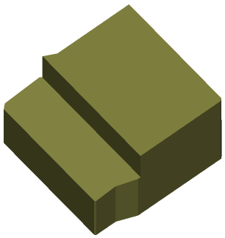}
        \includegraphics[width=0.85\linewidth]{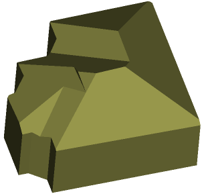}
        \includegraphics[width=\linewidth]{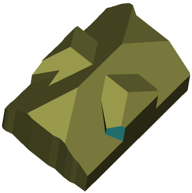}
        \includegraphics[width=0.9\linewidth]{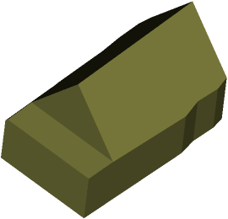}
        \includegraphics[width=0.95\linewidth]{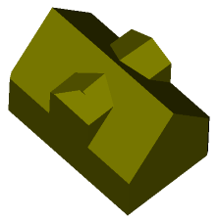}
        \includegraphics[width=0.8\linewidth]{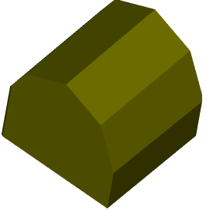}
        \includegraphics[width=0.8\linewidth]{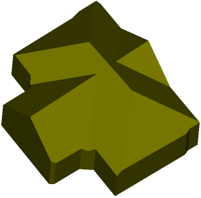}
        \includegraphics[width=0.8\linewidth]{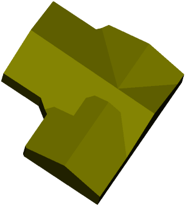}
        \includegraphics[width=0.85\linewidth]{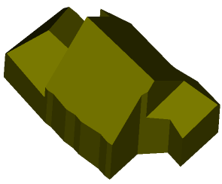}
        \subcaption*{\parbox{\linewidth}{\centering City3D\cite{huang2022city3d}\\PolyFit\cite{nan2017polyfit}}}
    \end{minipage}
    \hfill
    \begin{minipage}[t]{0.08\linewidth}
    \vspace{0pt}
        \includegraphics[width=1.1\linewidth]{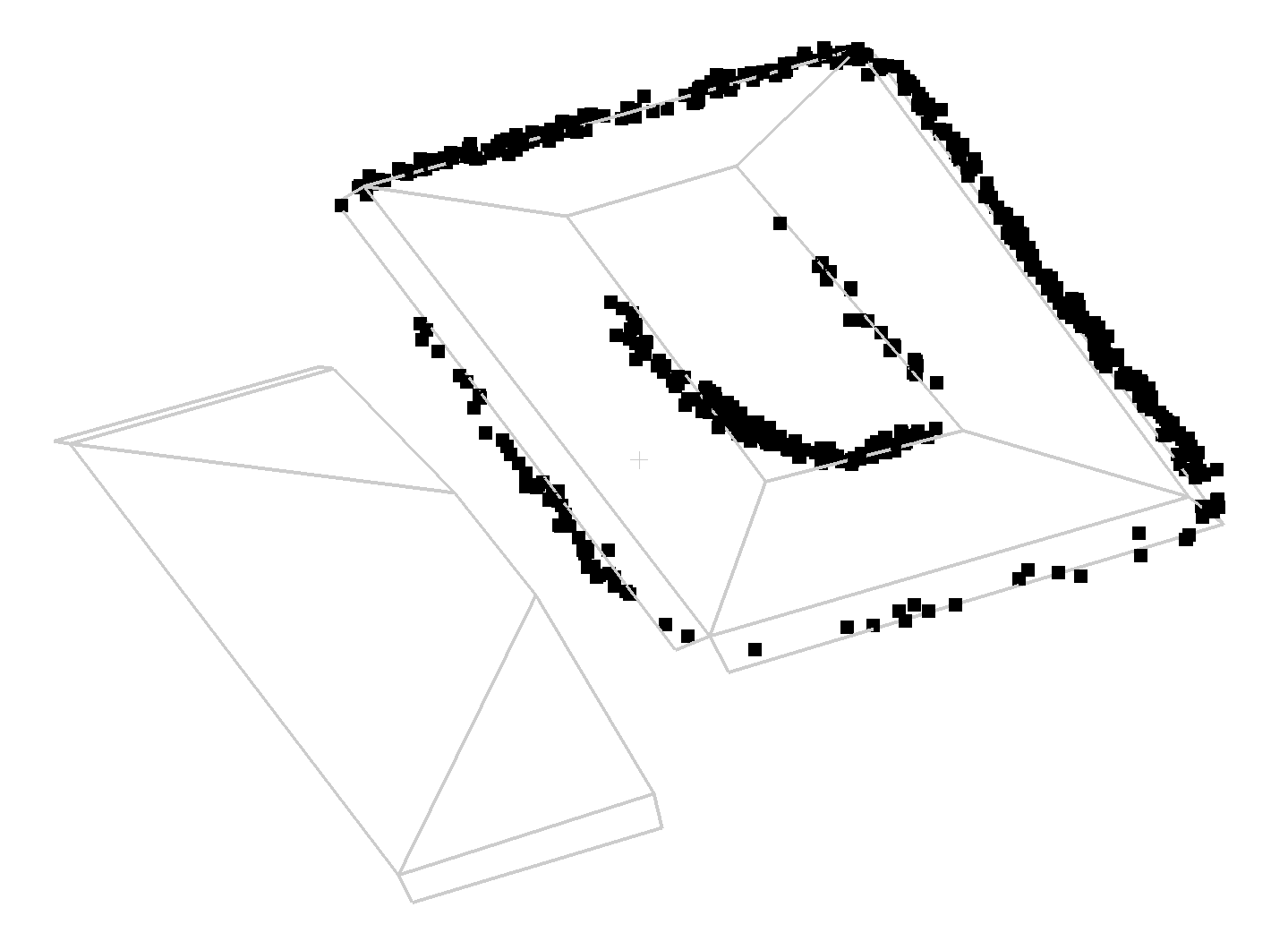}
        \includegraphics[width=1.1\linewidth]{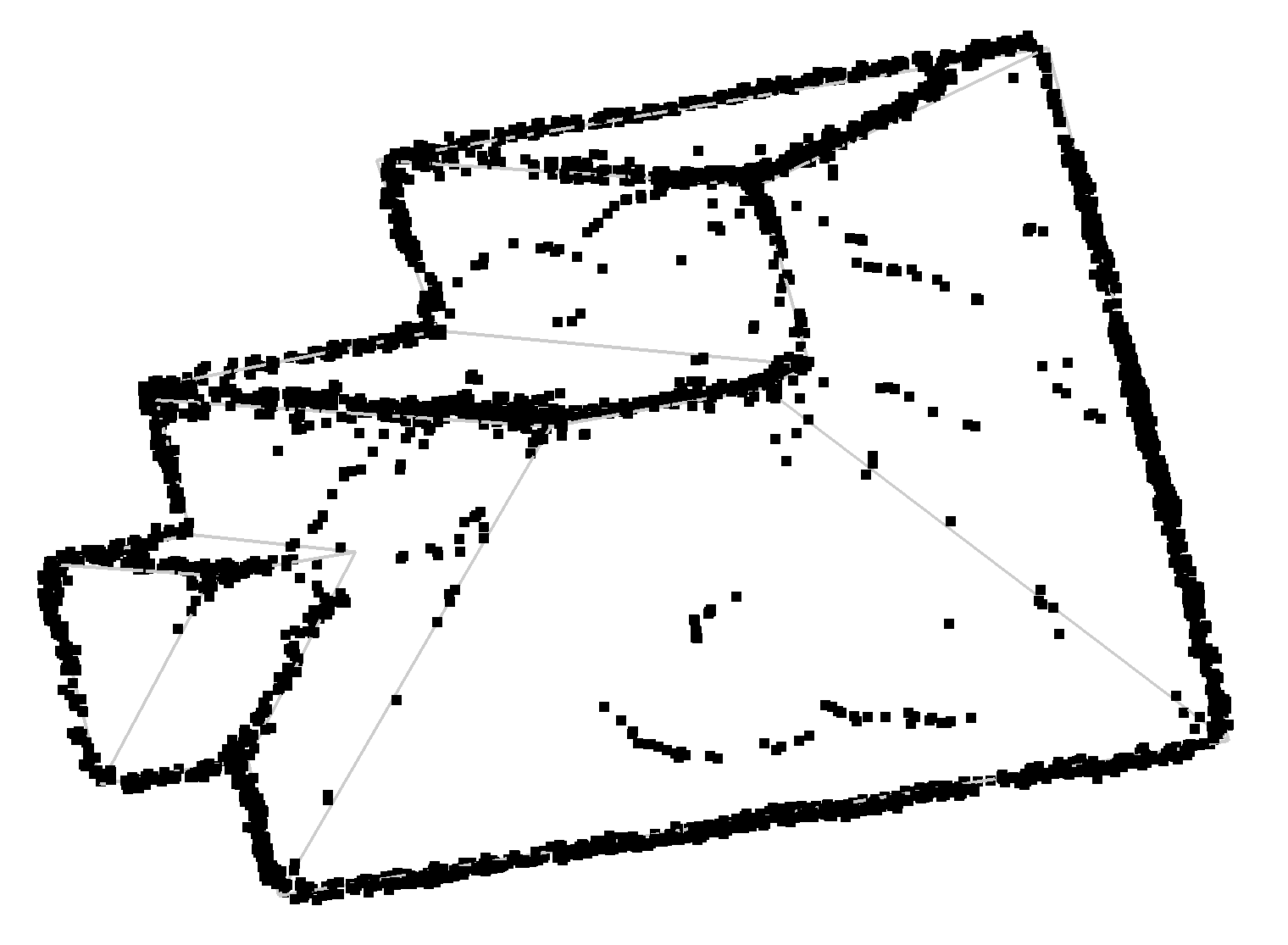}
        \includegraphics[width=1\linewidth]{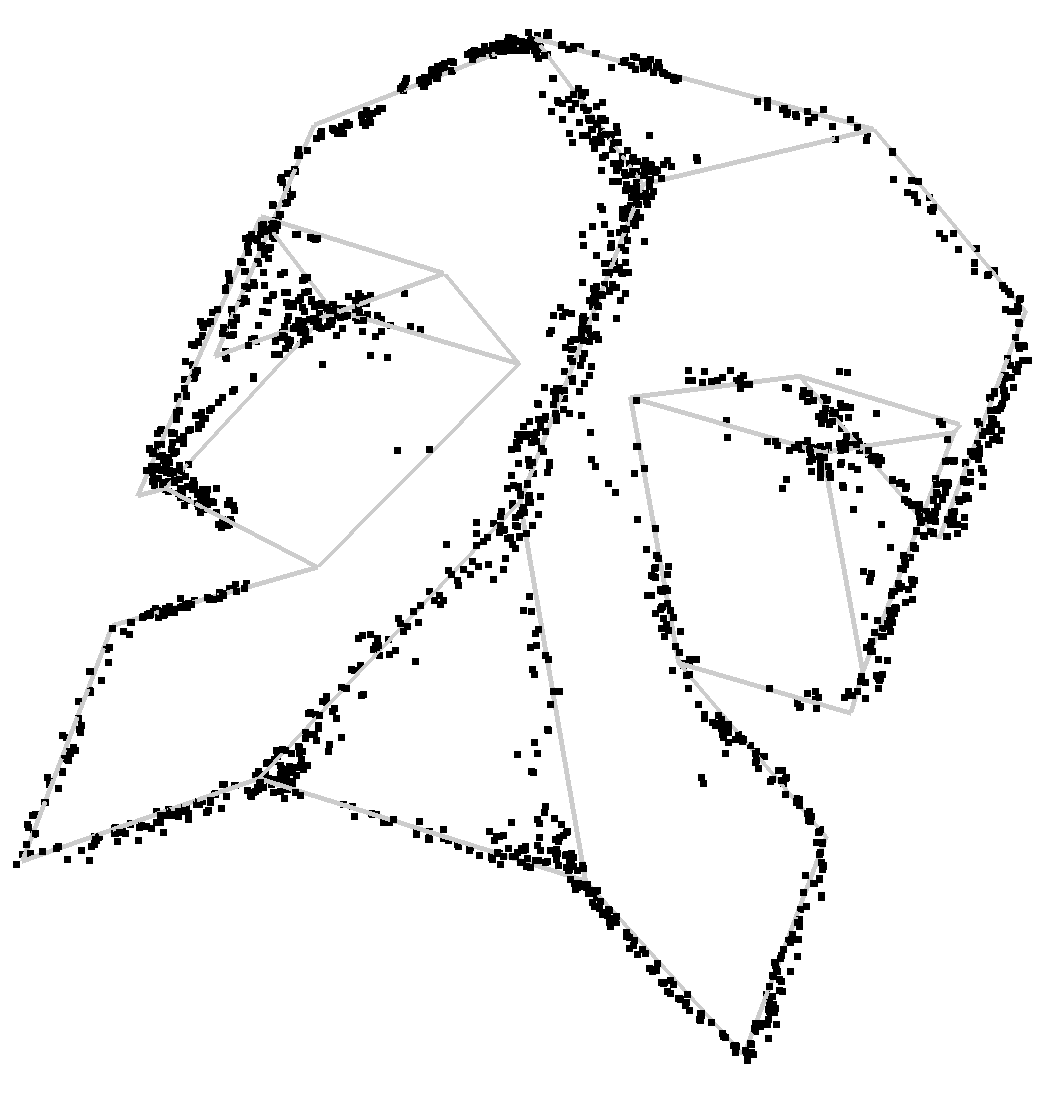}
        \includegraphics[width=0.9\linewidth]{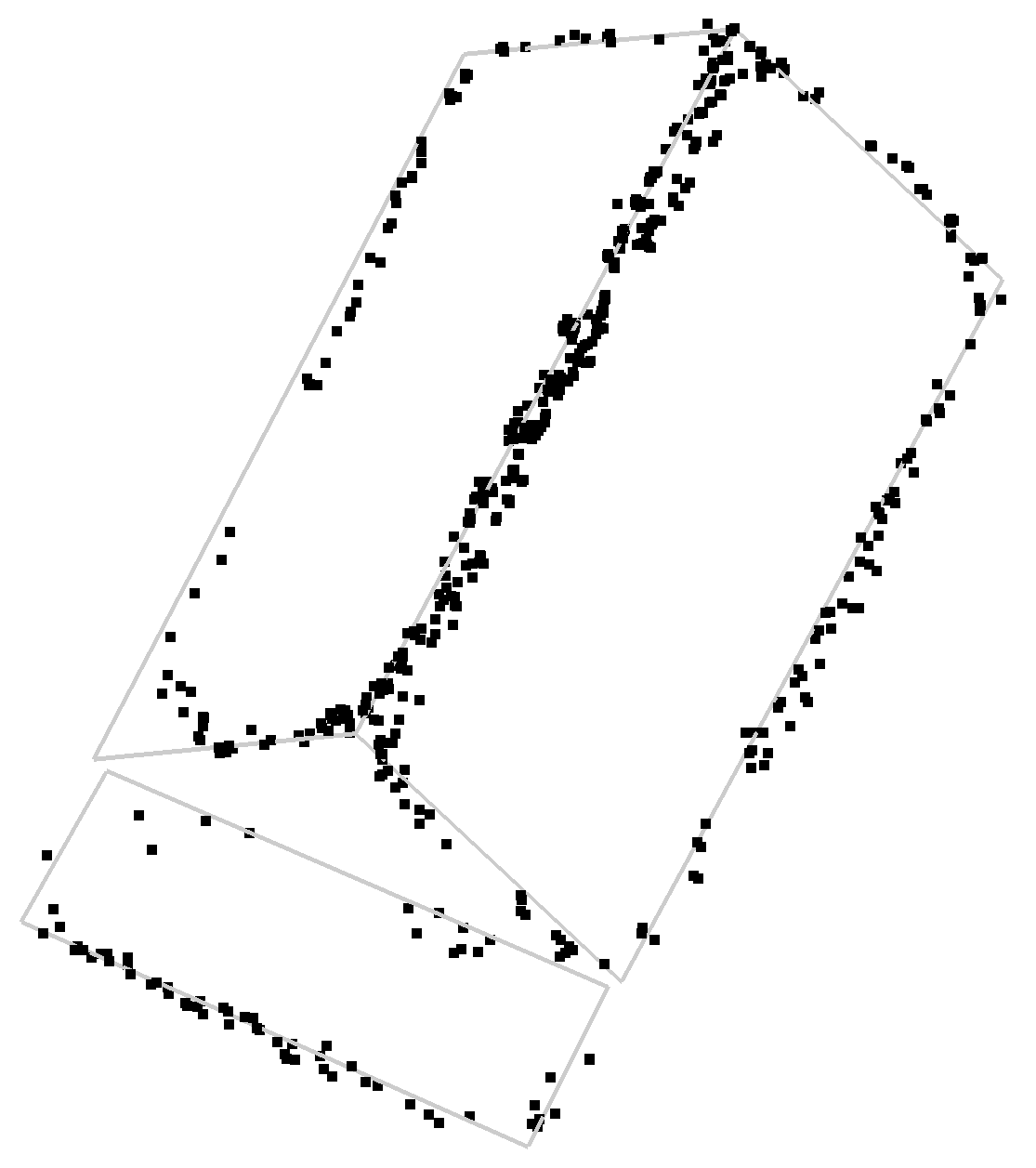}
        \includegraphics[width=\linewidth]{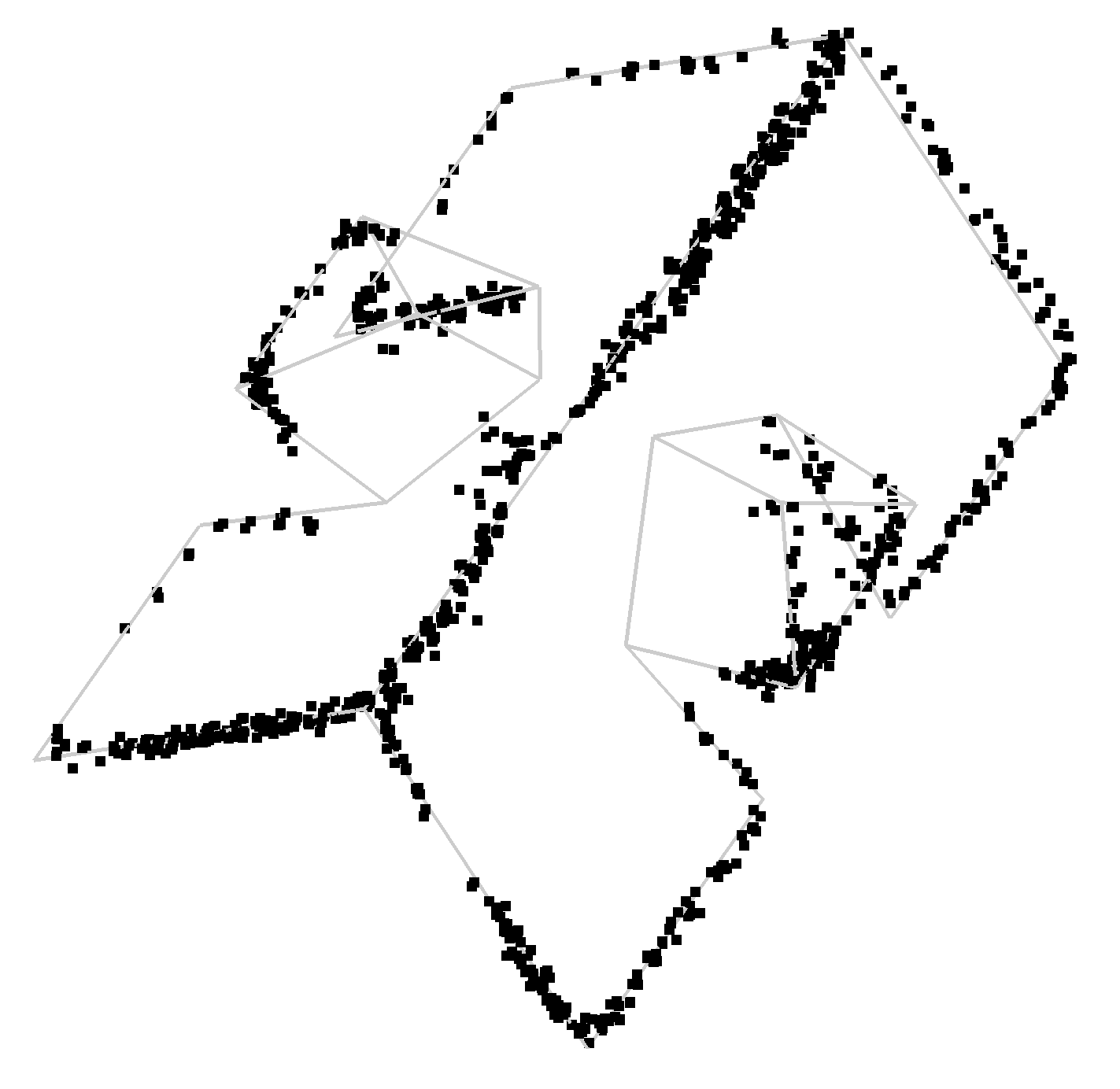}
        \includegraphics[width=\linewidth]{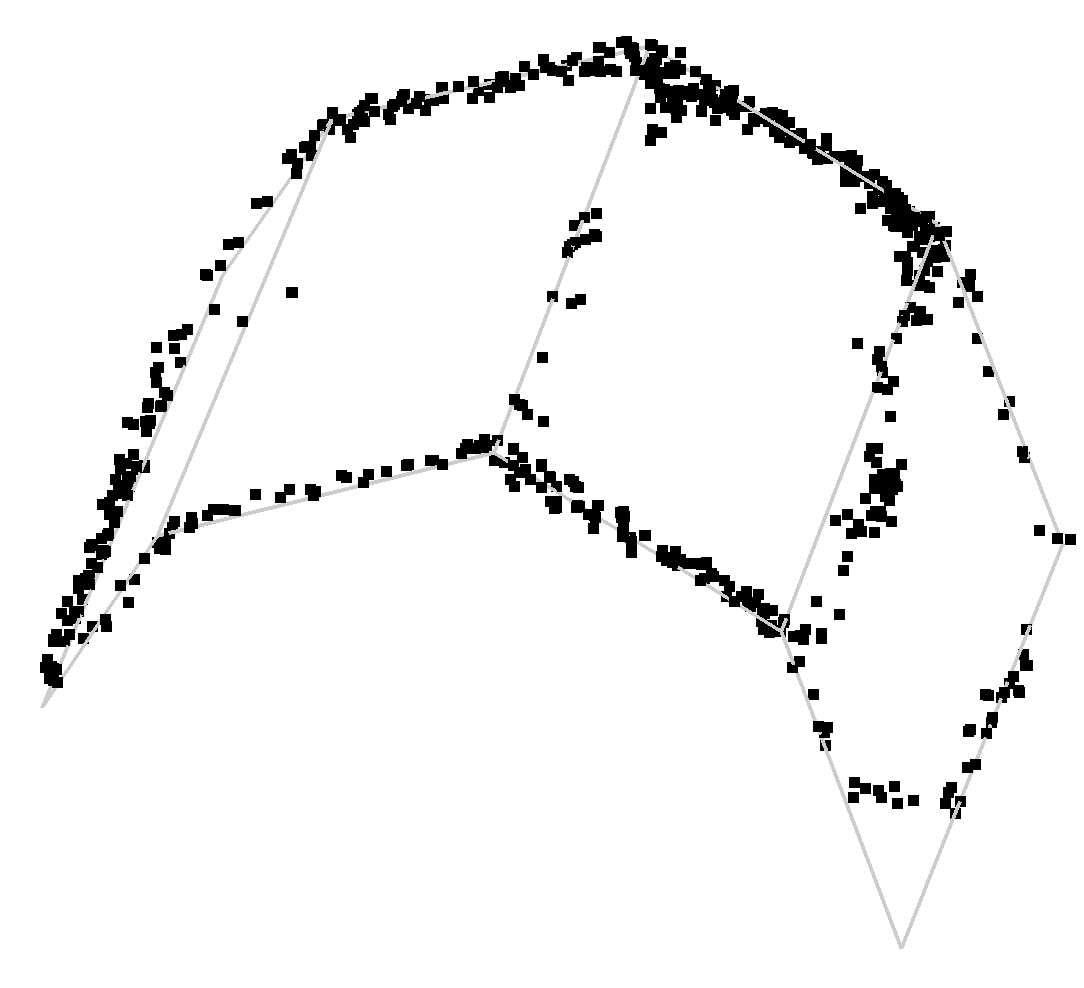}
        \includegraphics[width=\linewidth]{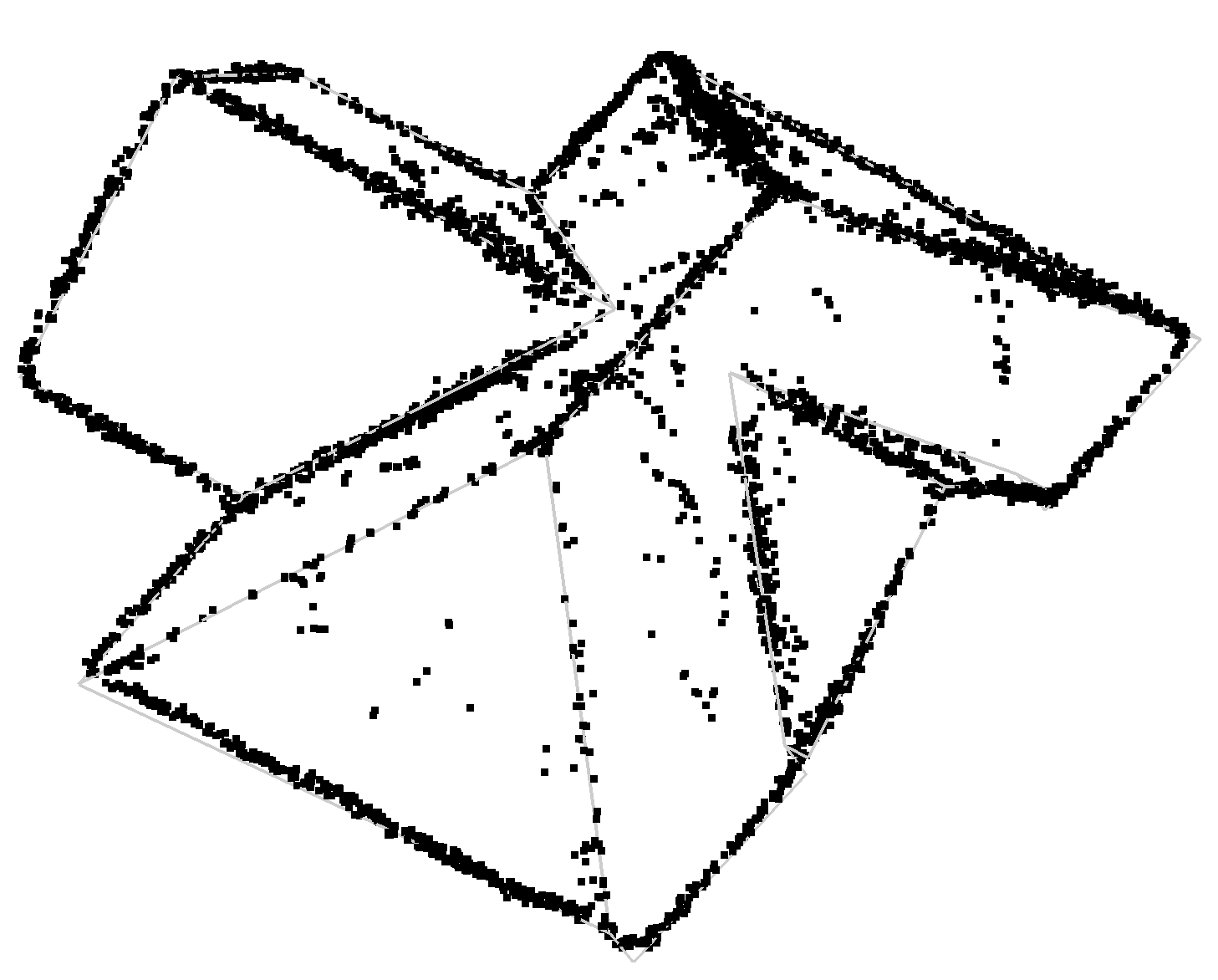}
        \includegraphics[width=\linewidth]{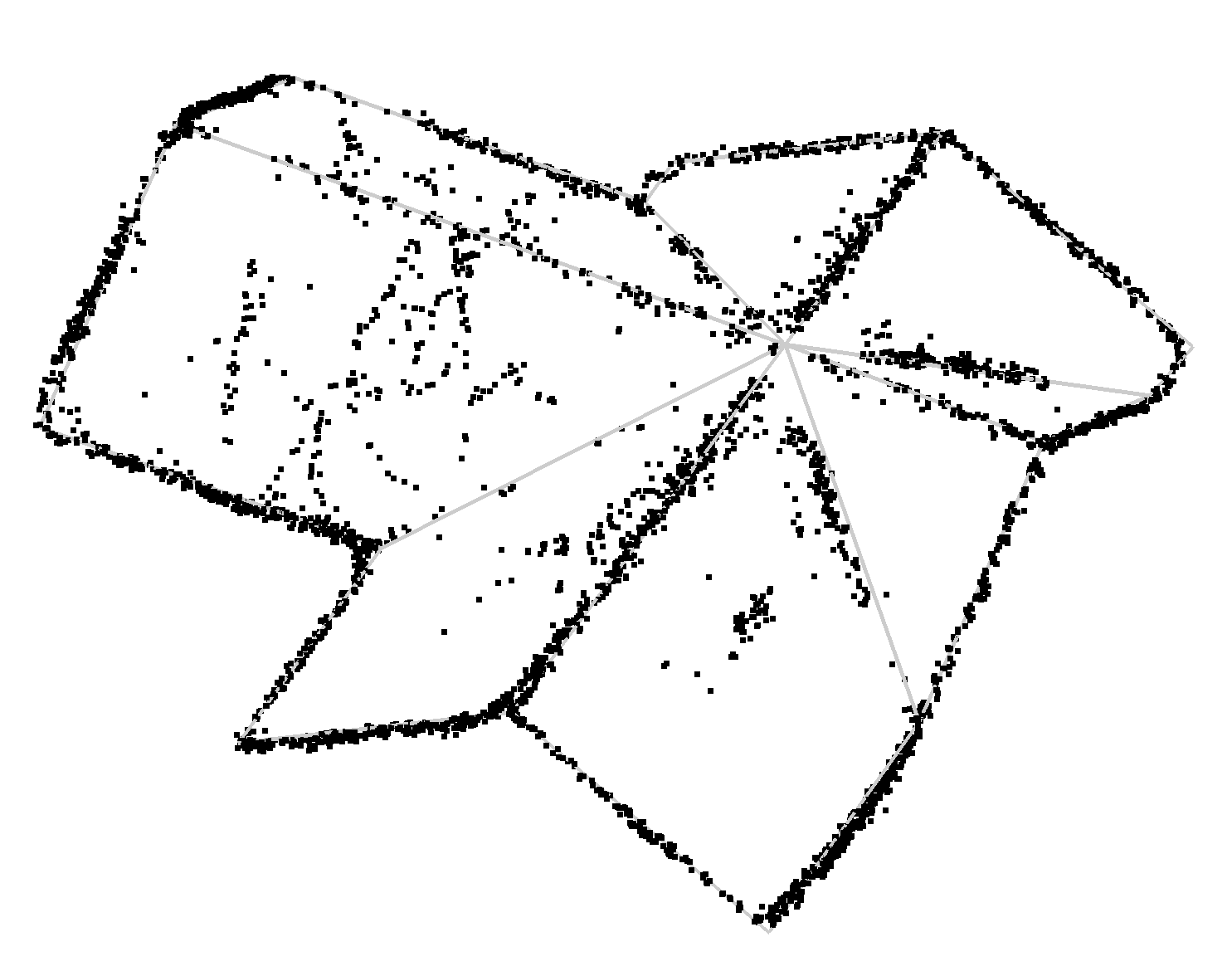}
        \includegraphics[width=\linewidth]{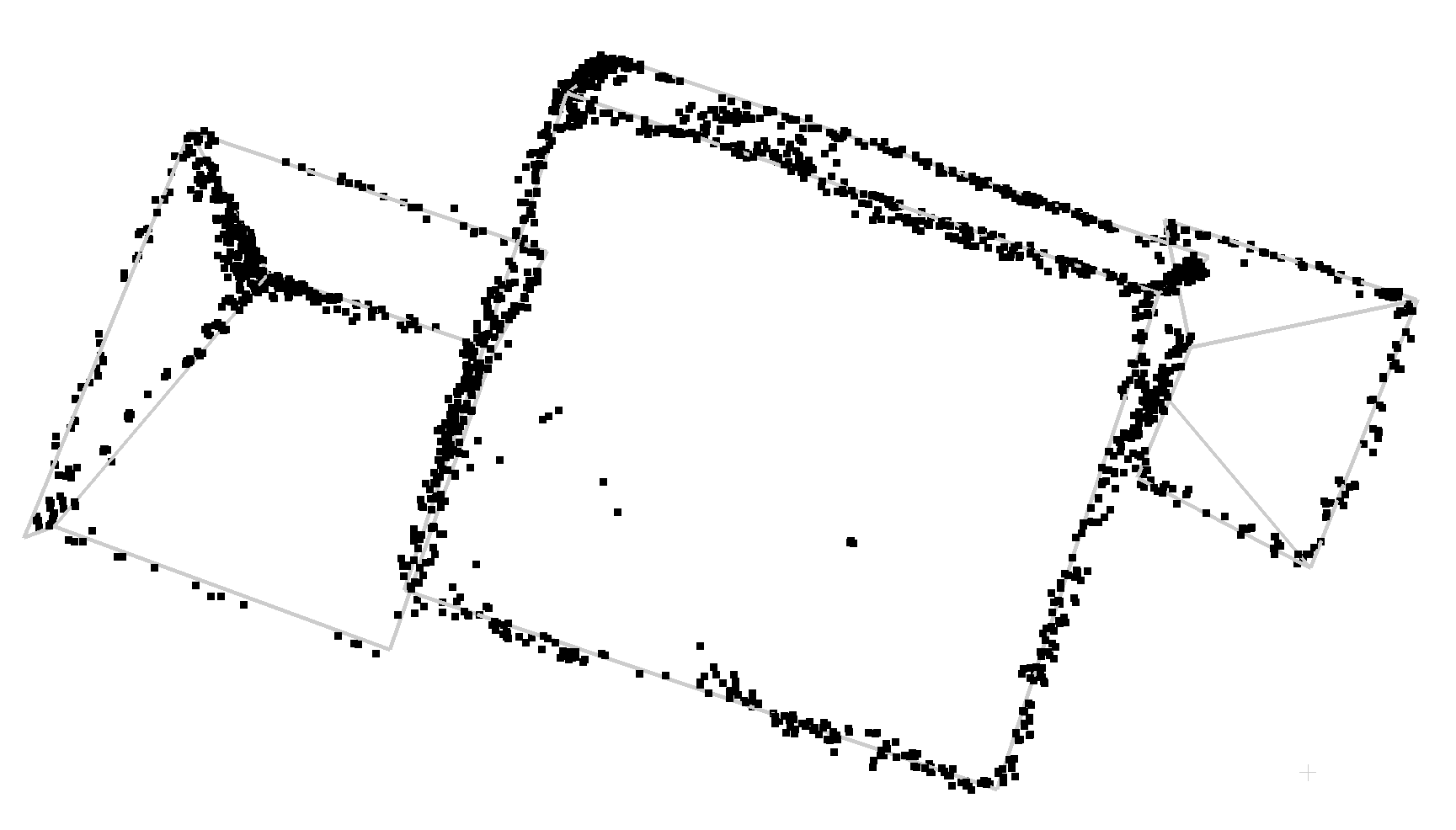}
        \subcaption*{\parbox{\linewidth}{\centering EC-Net\cite{yu2018ec}}}
    \end{minipage}
    \hfill
    \begin{minipage}[t]{0.08\linewidth}
    \vspace{0pt}
        \includegraphics[width=\linewidth]{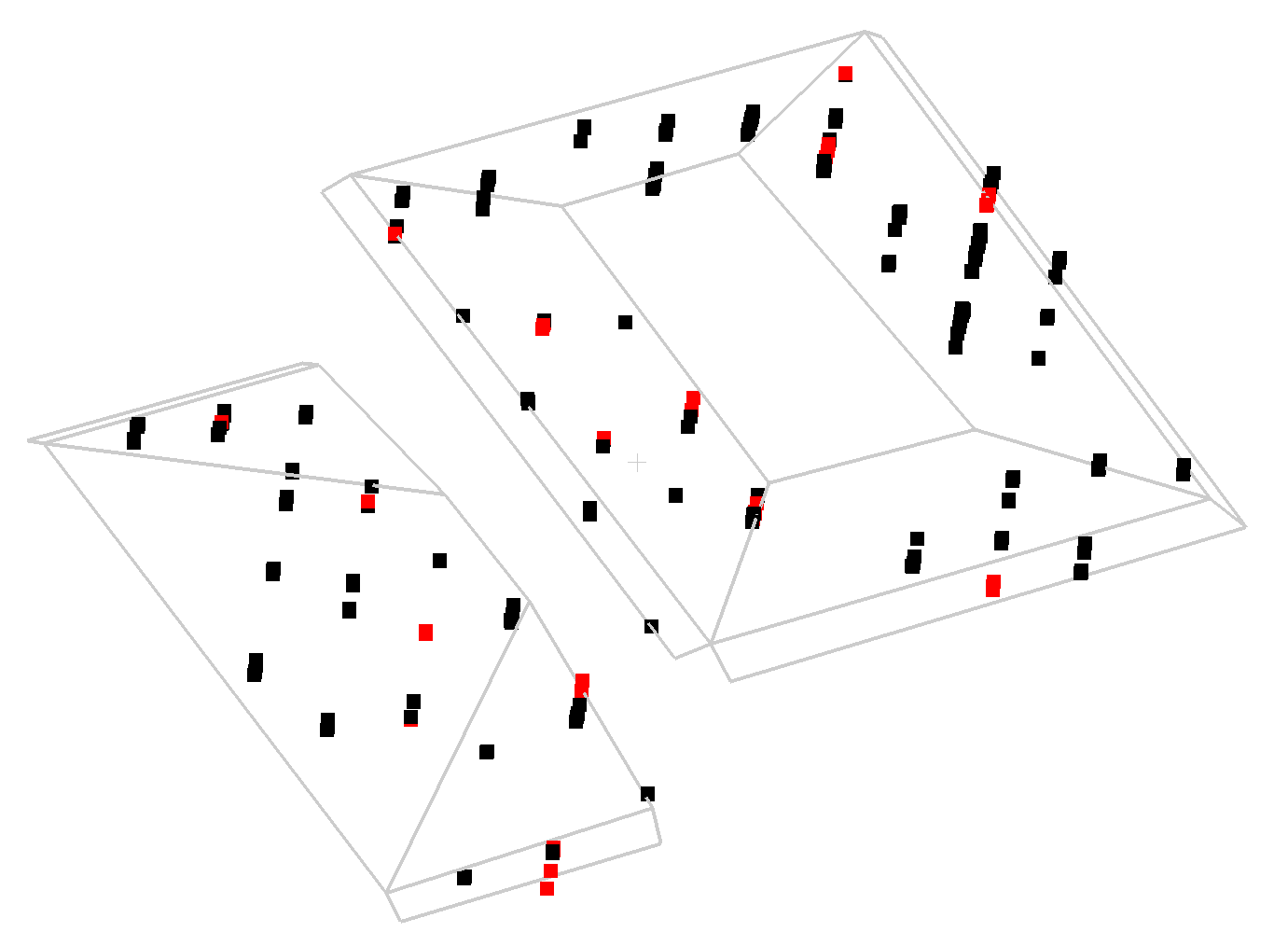}
        \includegraphics[width=1.1\linewidth]{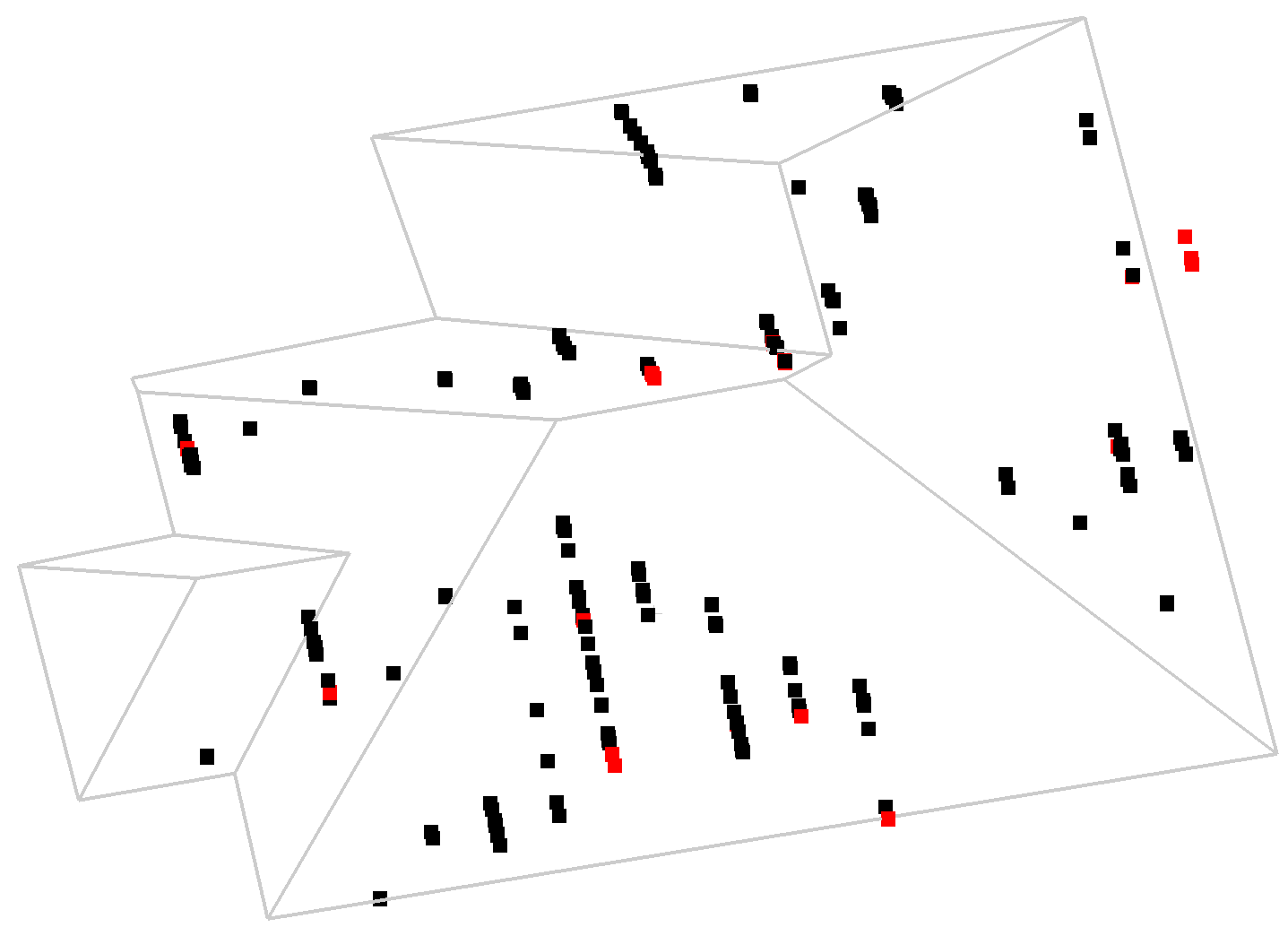}
        \includegraphics[width=1.1\linewidth]{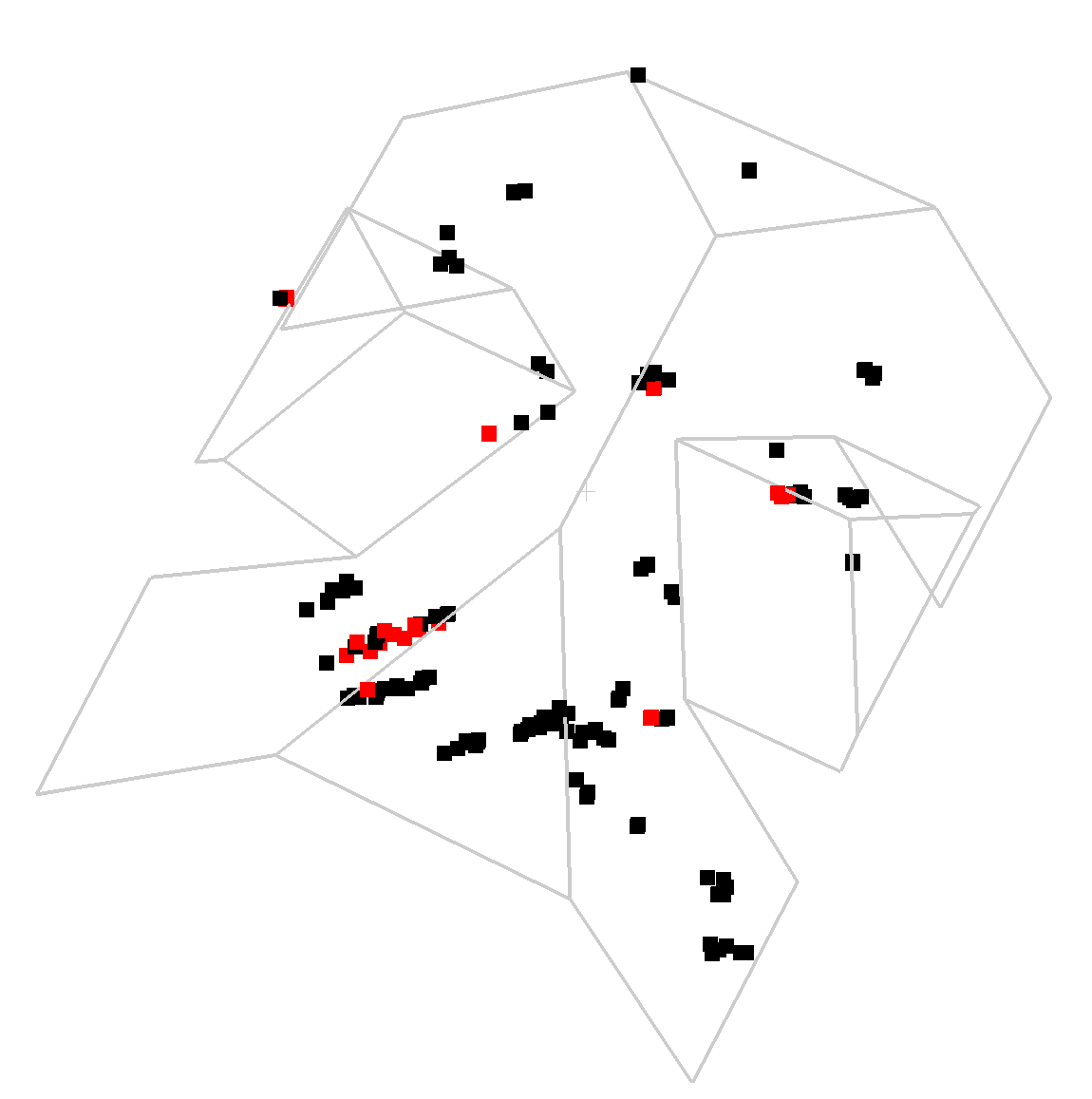}
        \includegraphics[width=1\linewidth]{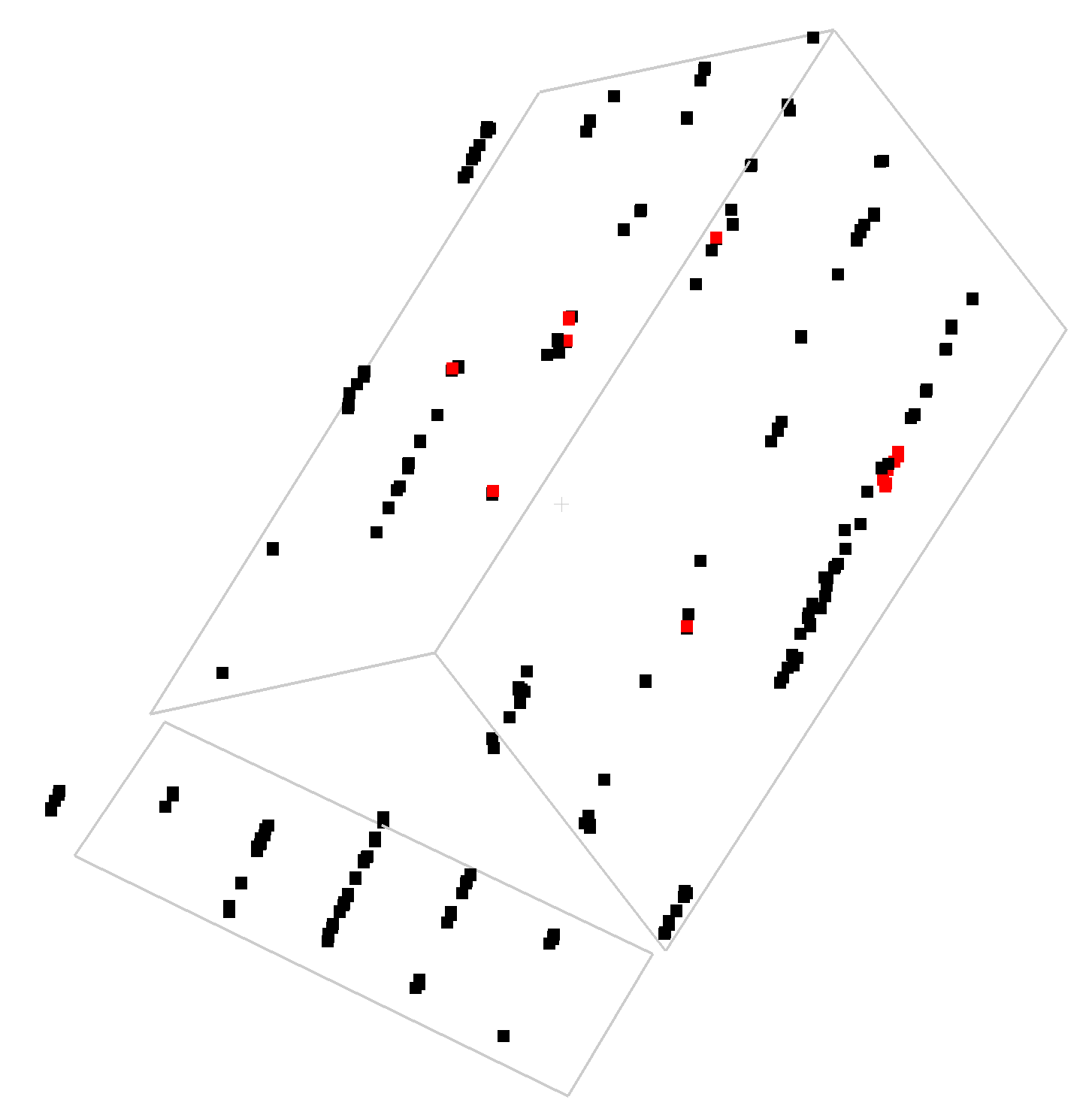}
        \includegraphics[width=\linewidth]{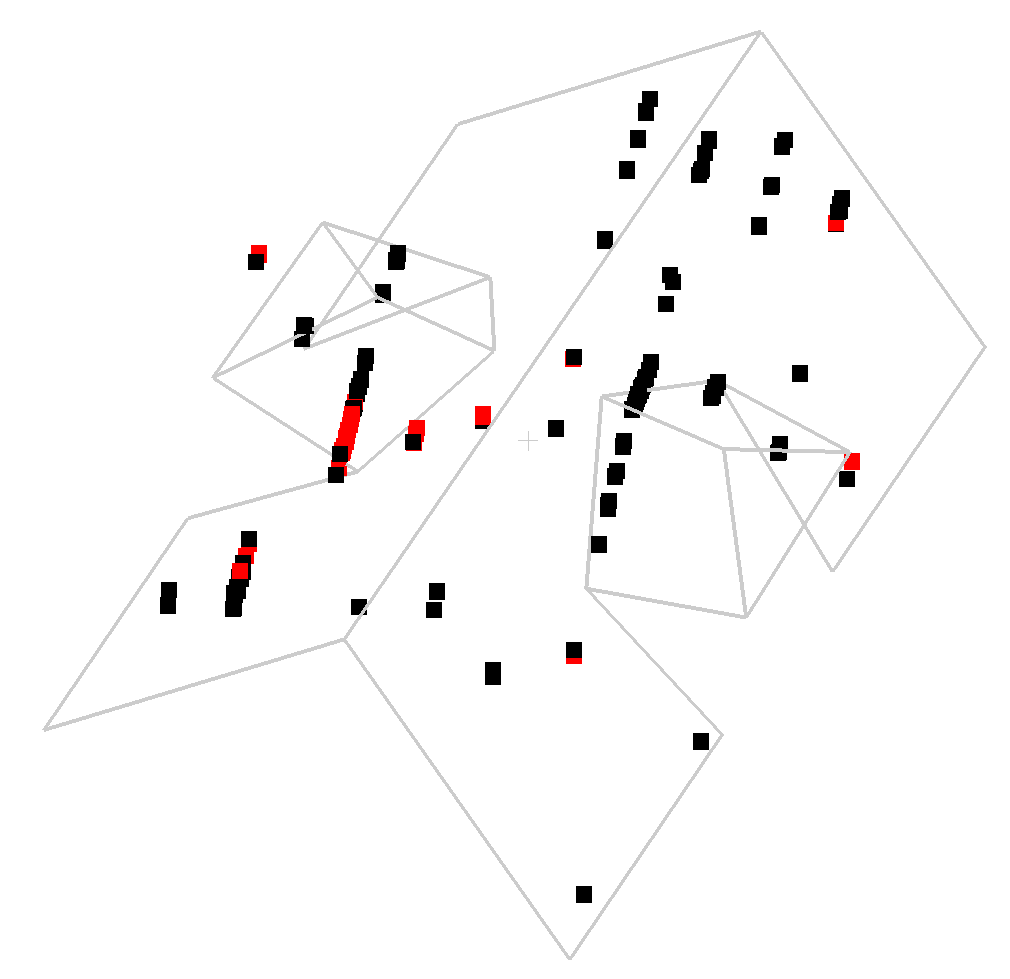}
        \includegraphics[width=\linewidth]{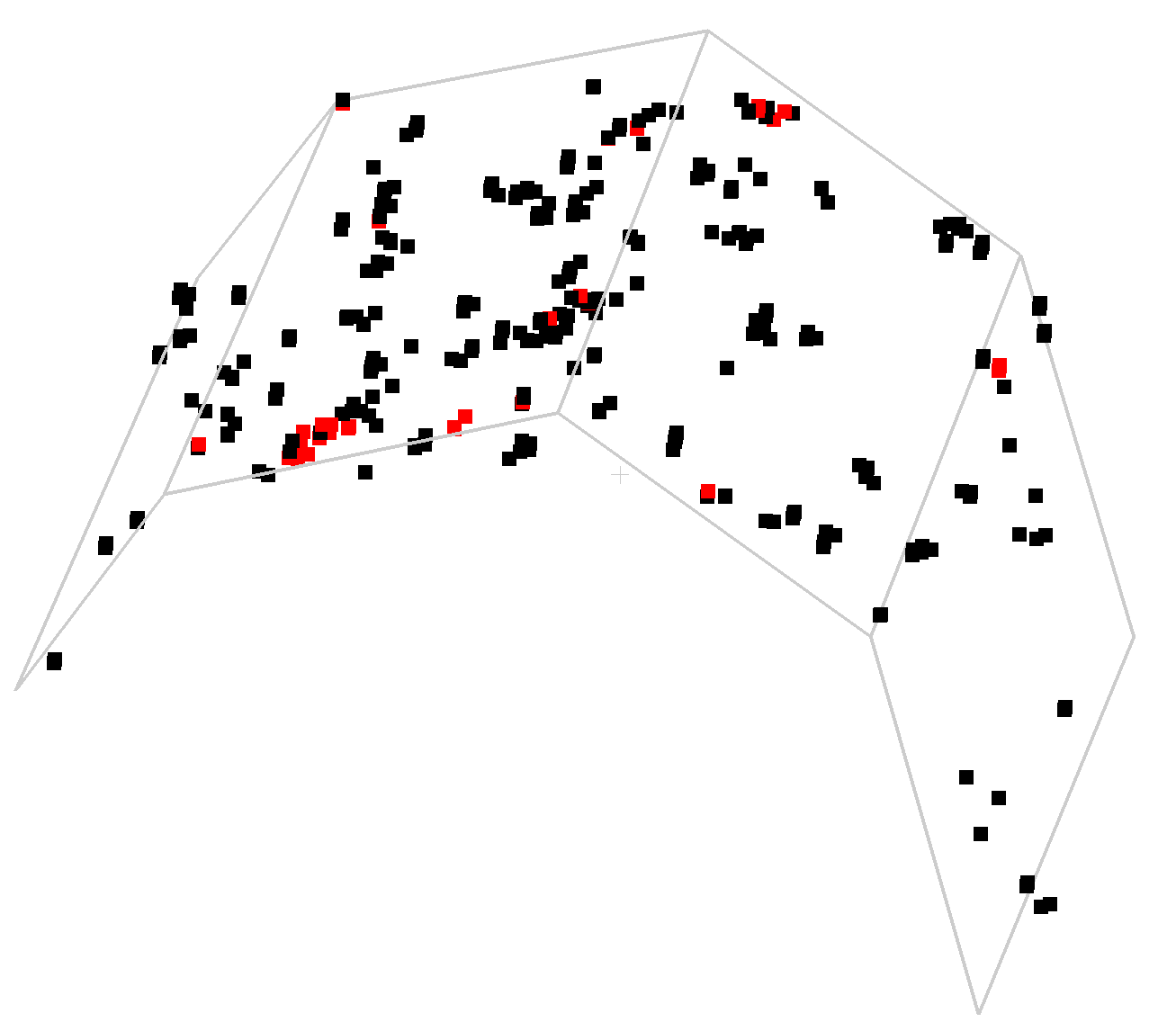}
        \includegraphics[width=\linewidth]{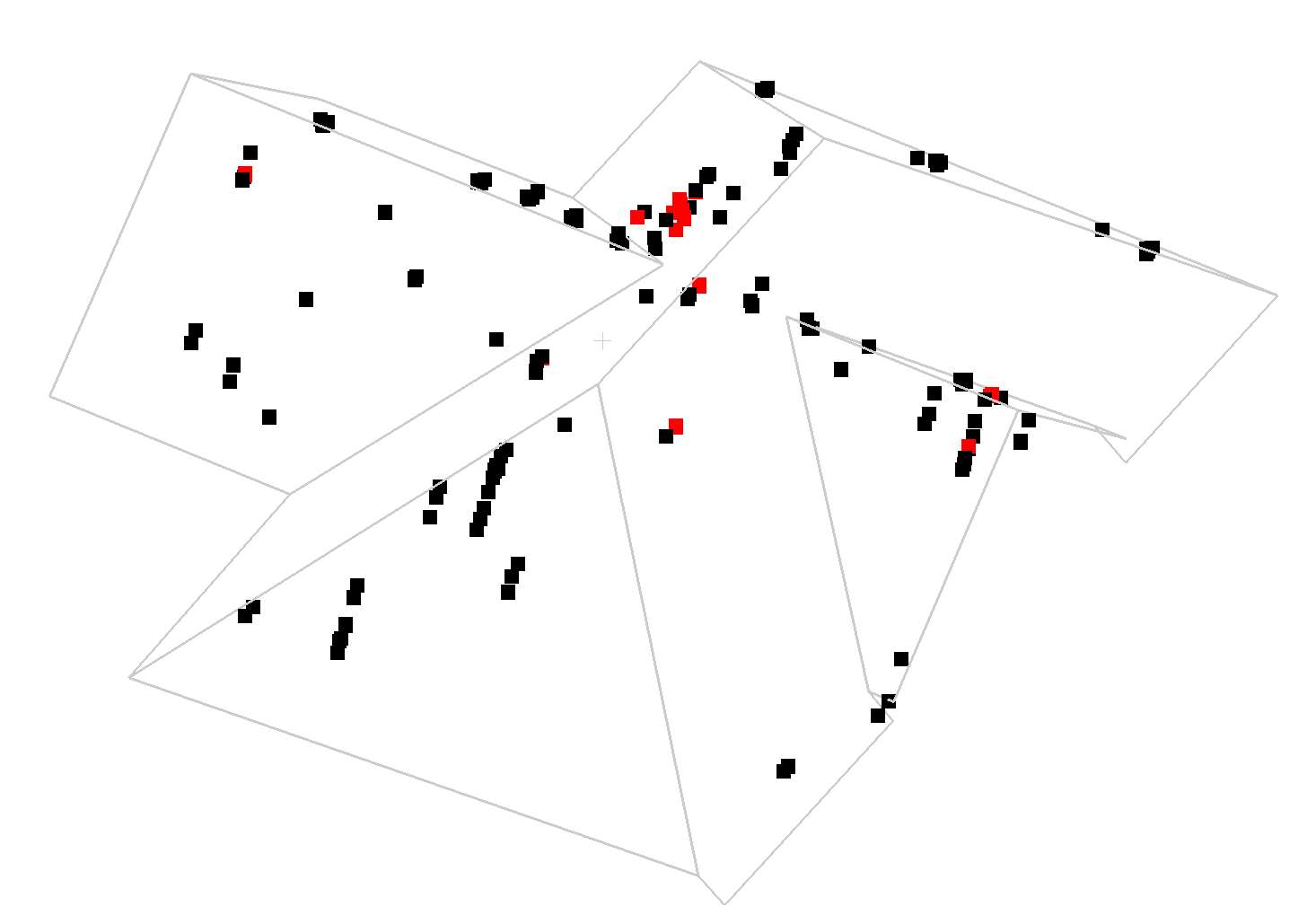}
        \includegraphics[width=\linewidth]{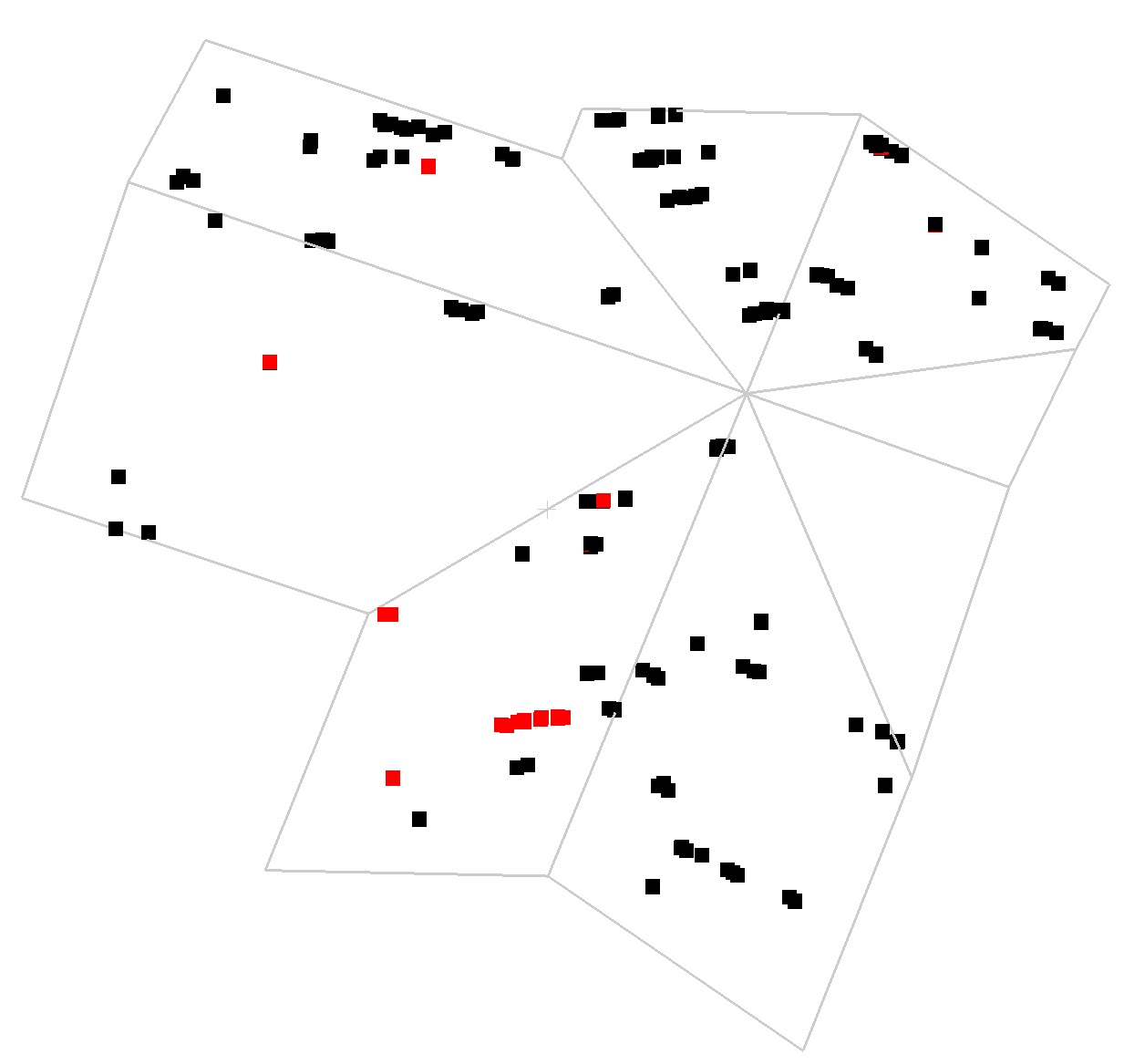}
        \includegraphics[width=\linewidth]{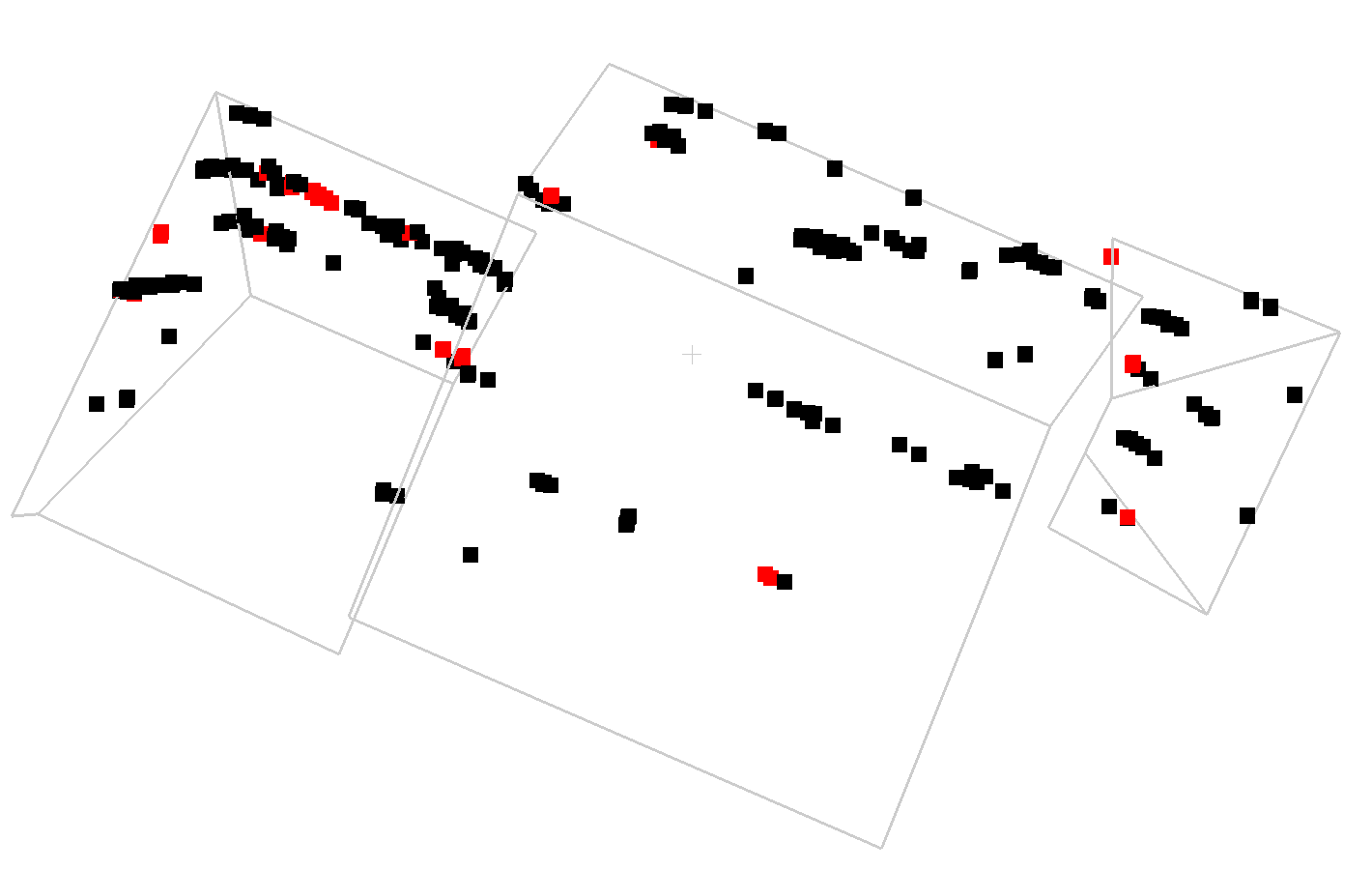}
        \subcaption*{\parbox{\linewidth}{\centering PIE-Net\cite{wang2020pie}}}
    \end{minipage}
    \hfill
    \begin{minipage}[t]{0.08\linewidth}
    \vspace{0pt}
        \includegraphics[width=1.1\linewidth]{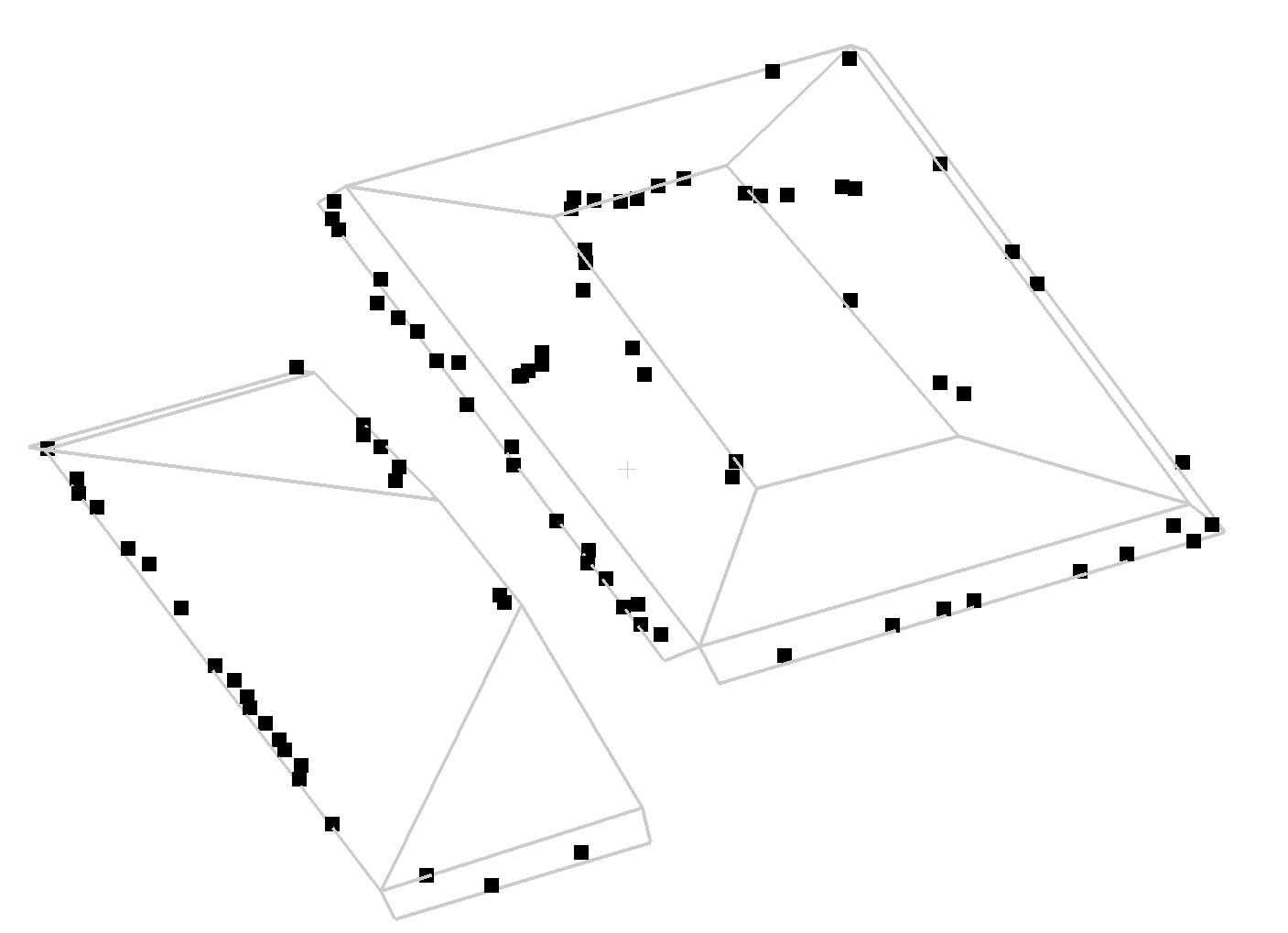}
        \includegraphics[width=1.1\linewidth]{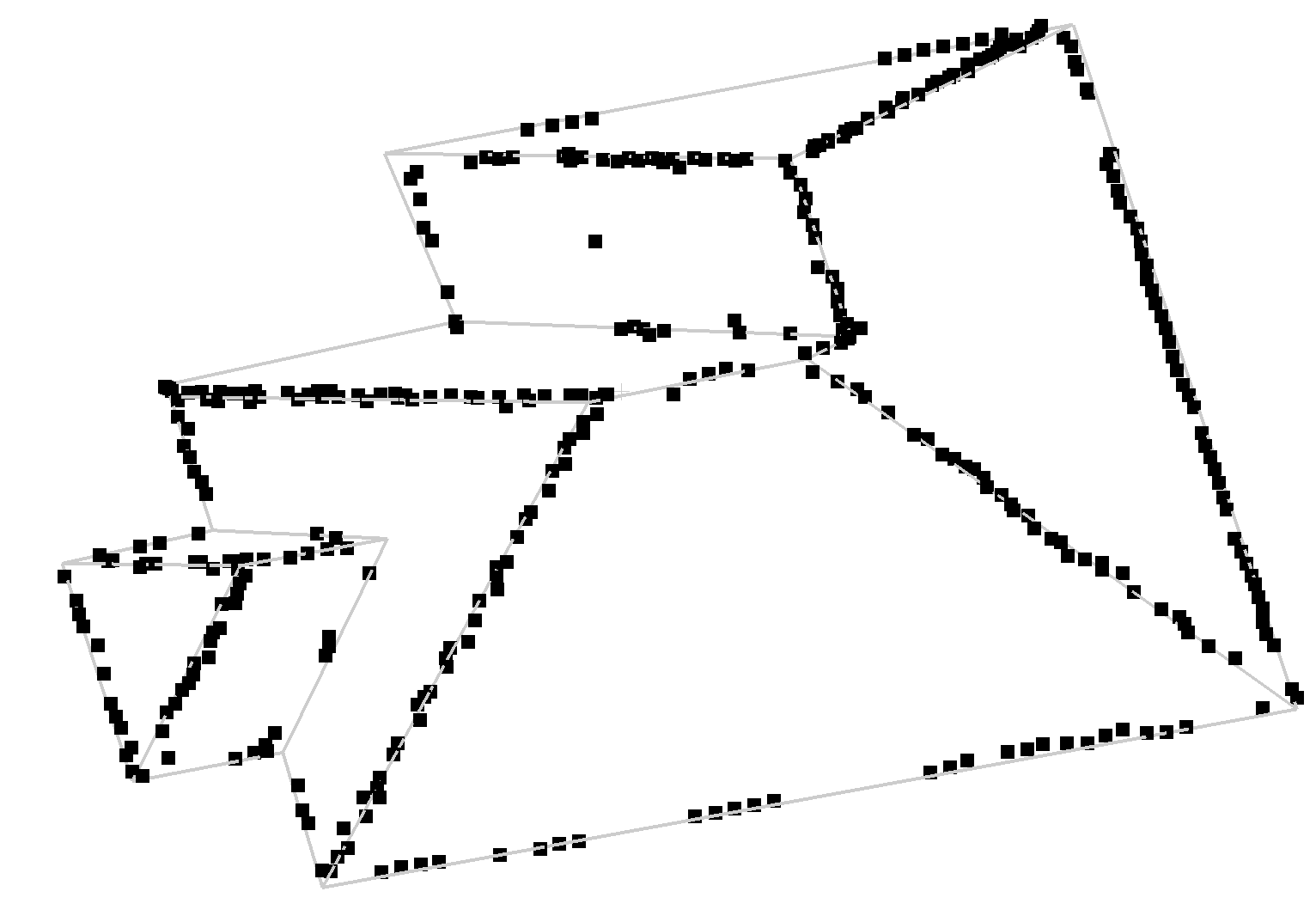}
        \includegraphics[width=1\linewidth]{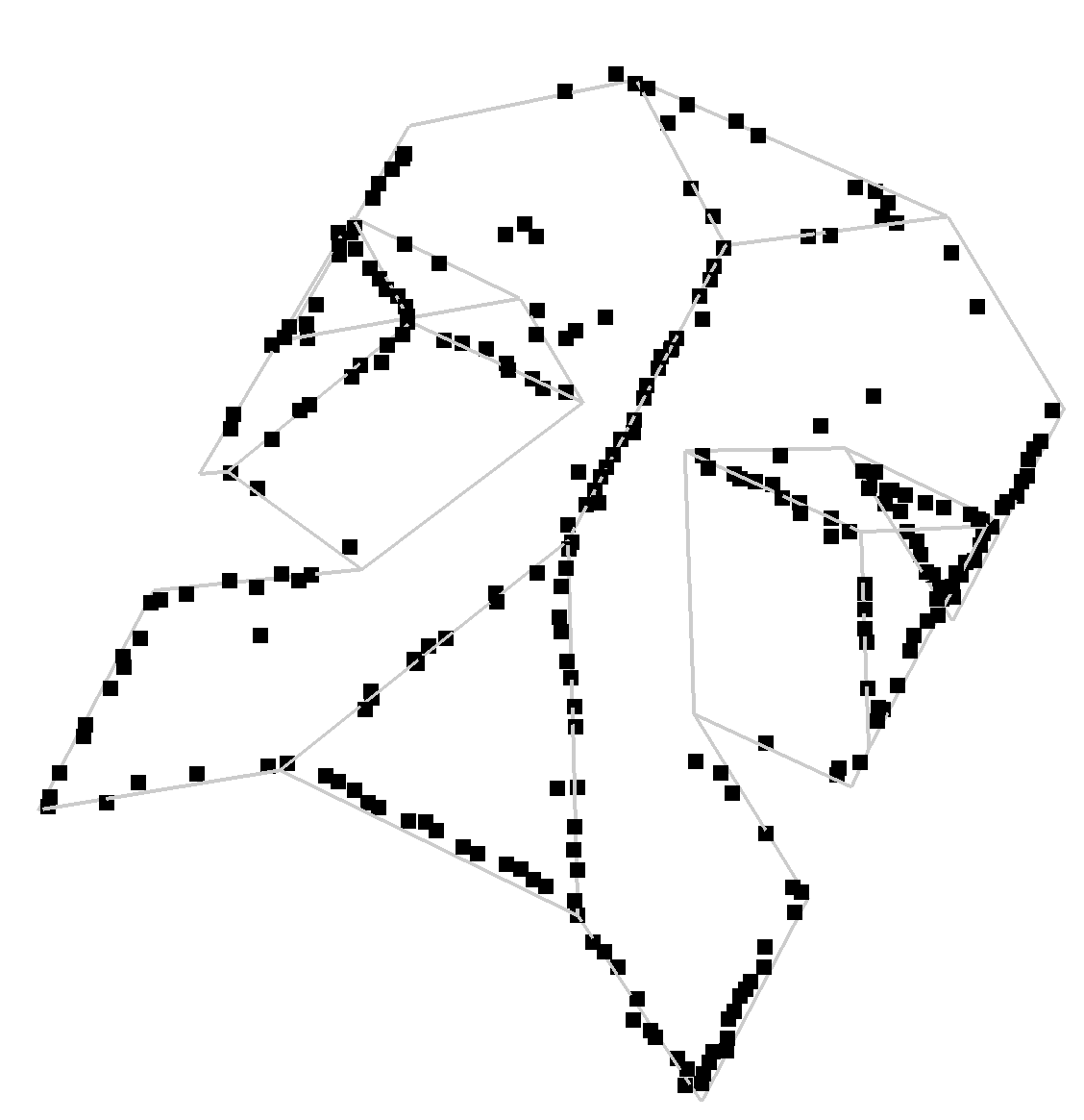}
        \includegraphics[width=1\linewidth]{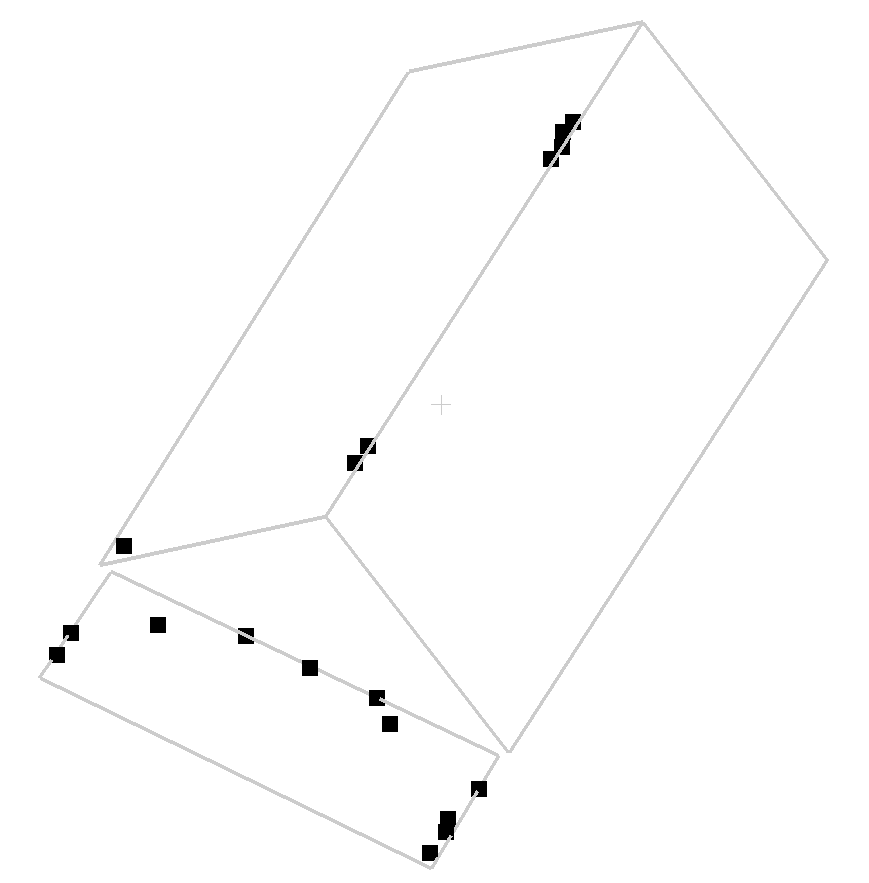}
        \includegraphics[width=\linewidth]{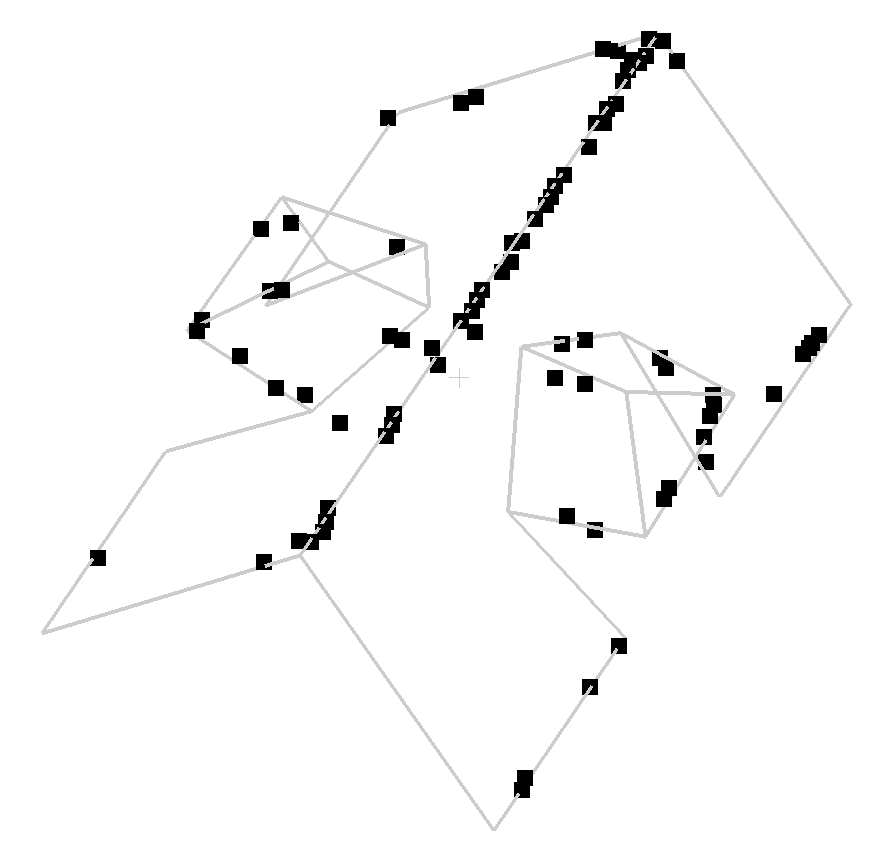}
        \includegraphics[width=\linewidth]{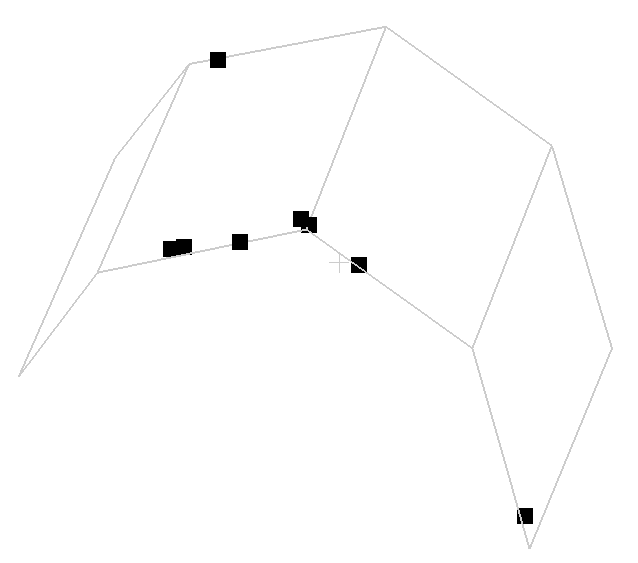}
        \includegraphics[width=\linewidth]{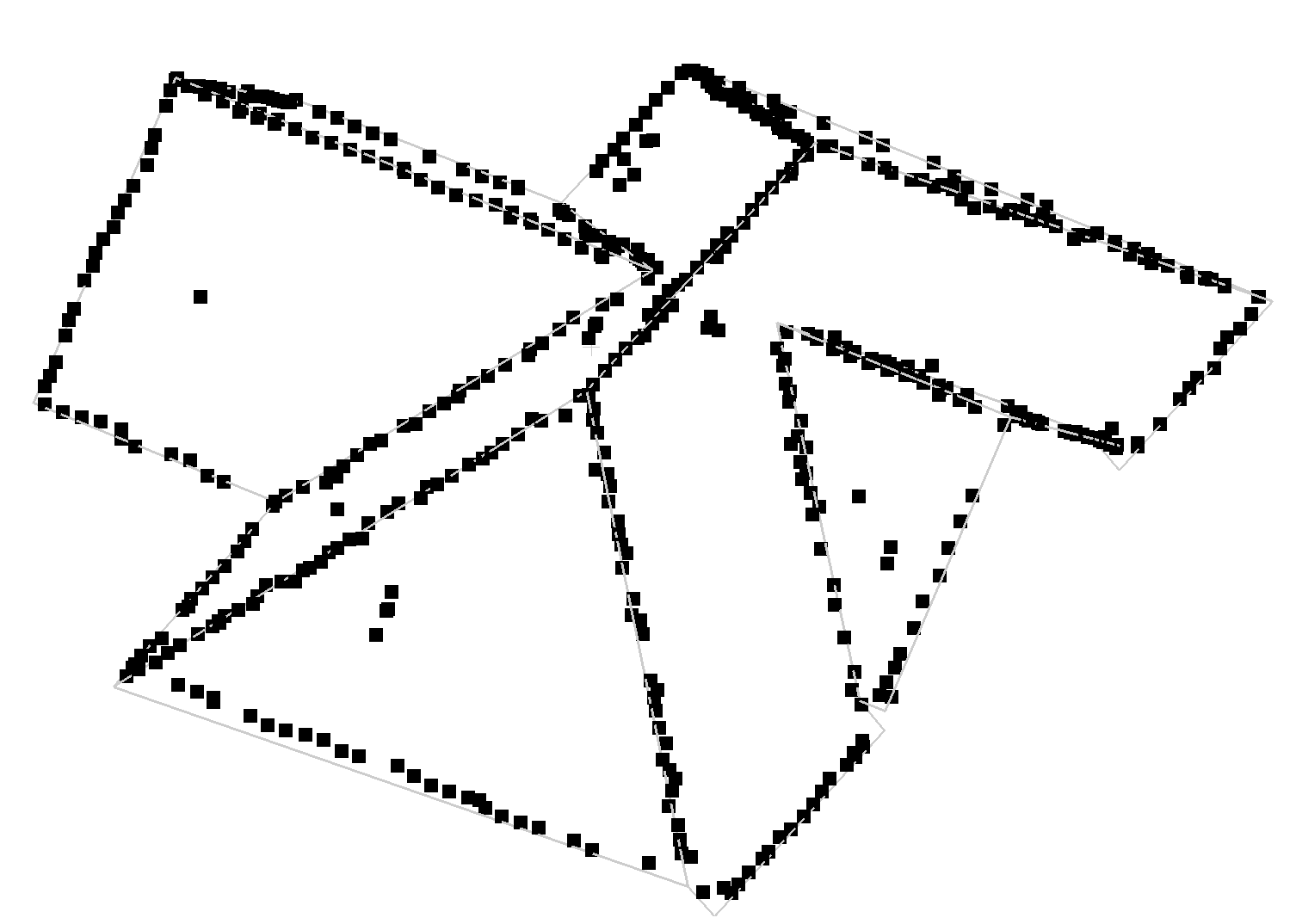}
        \includegraphics[width=\linewidth]{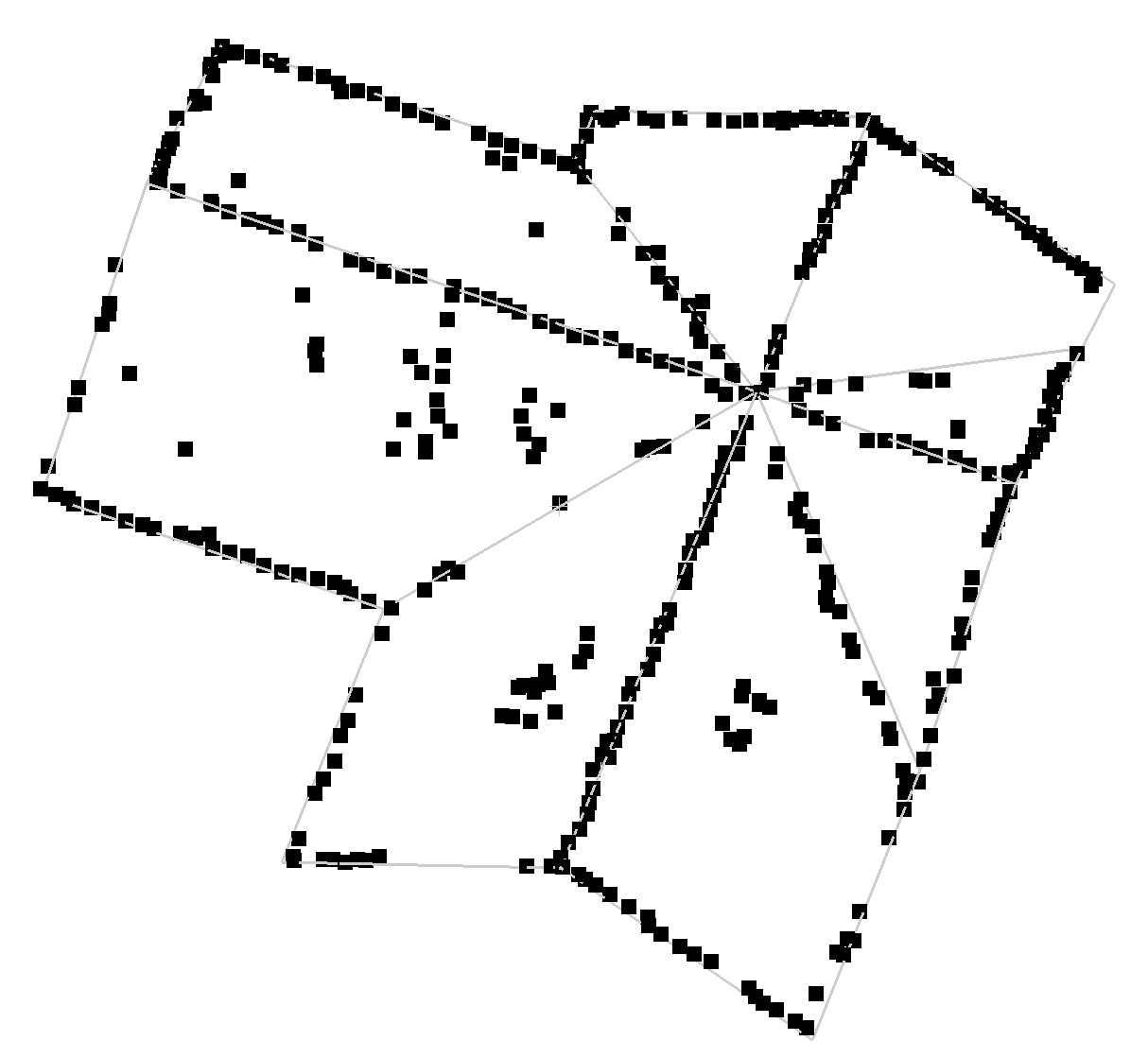}
        \includegraphics[width=\linewidth]{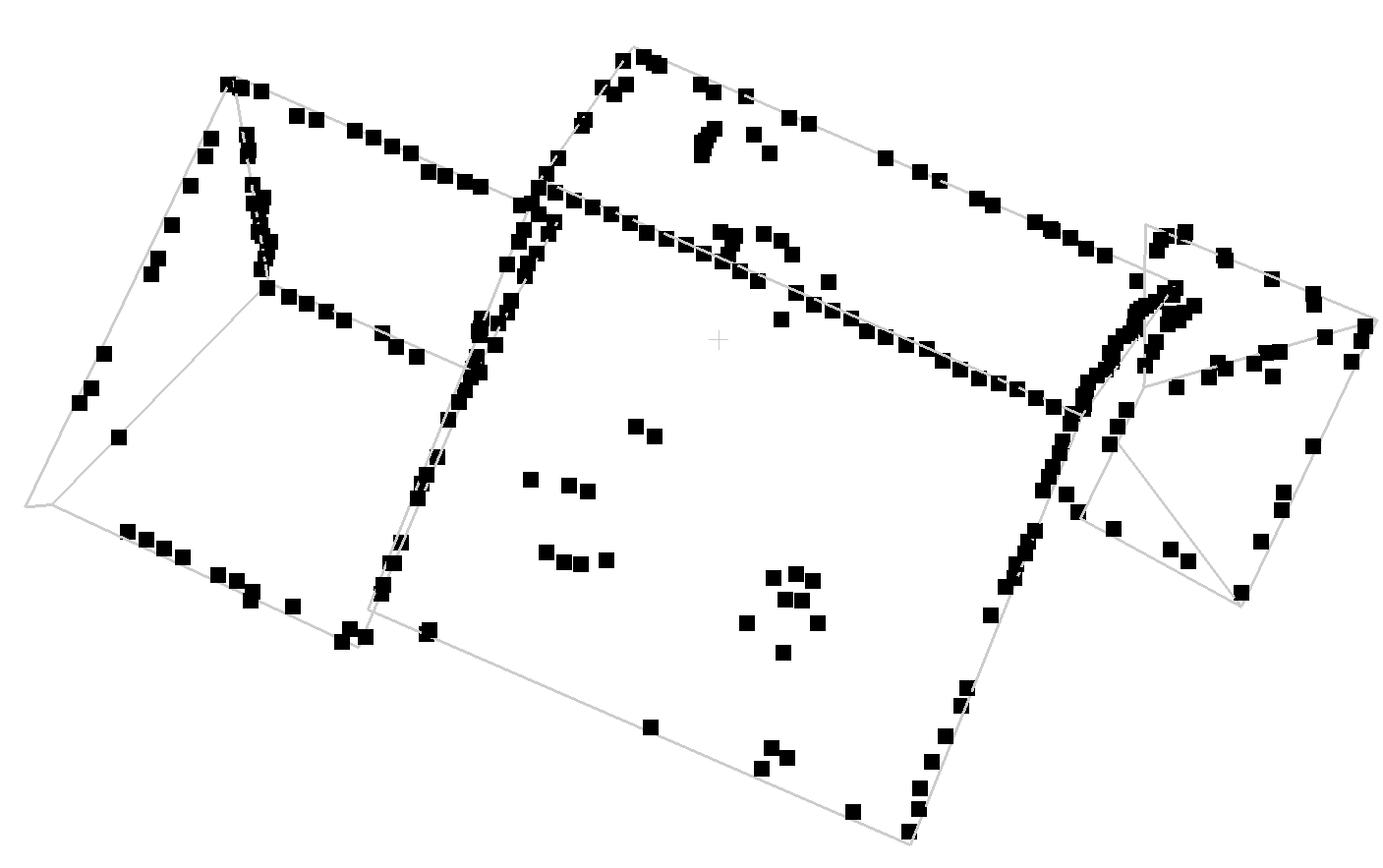}
        \subcaption*{\parbox{\linewidth}{\centering NerVE\cite{zhu2023nerve}}}
    \end{minipage}
    \hfill
    \begin{minipage}[t]{0.08\linewidth}
    \vspace{0pt}
        \includegraphics[width=1.1\linewidth]{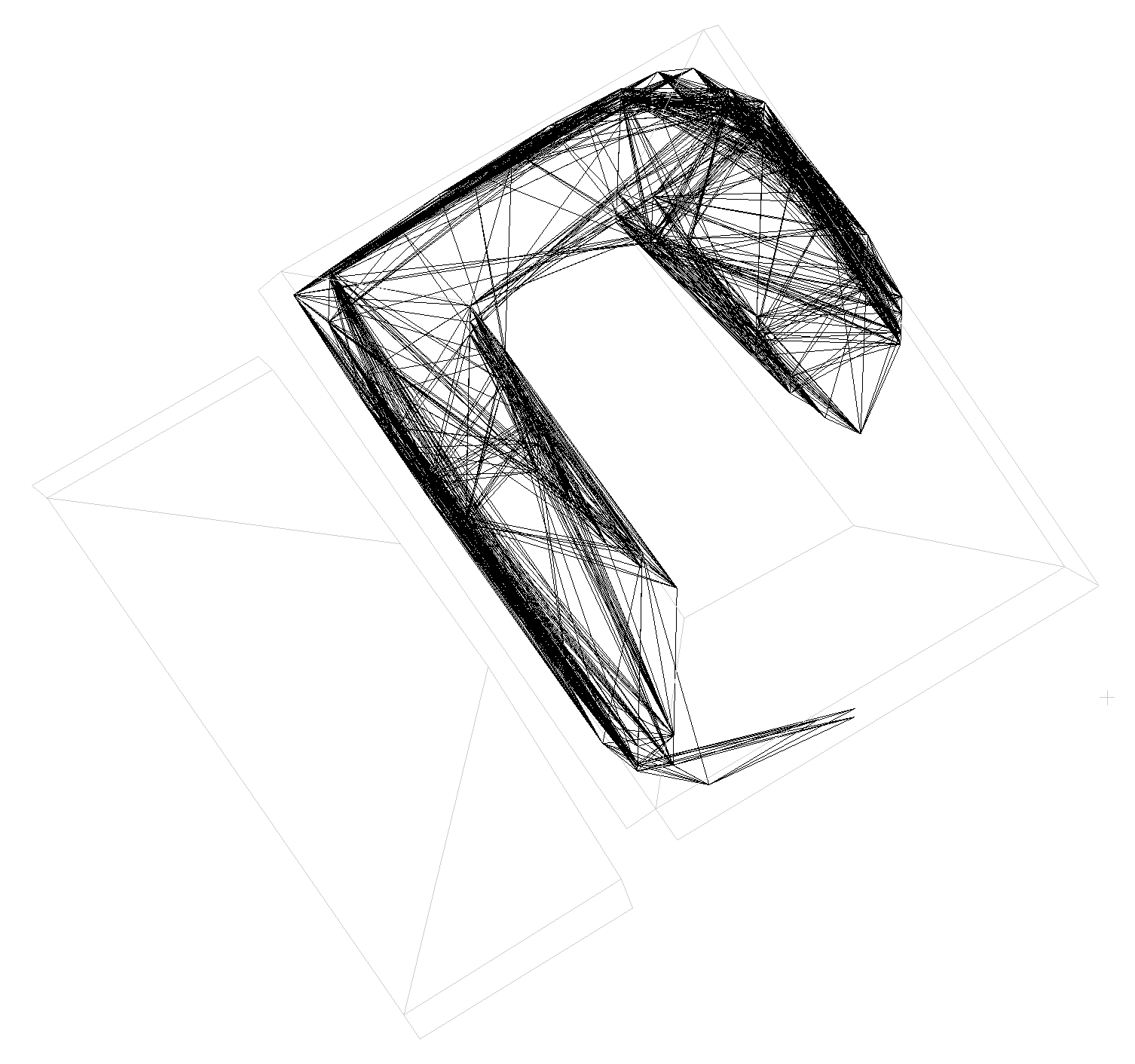}
        \includegraphics[width=\linewidth]{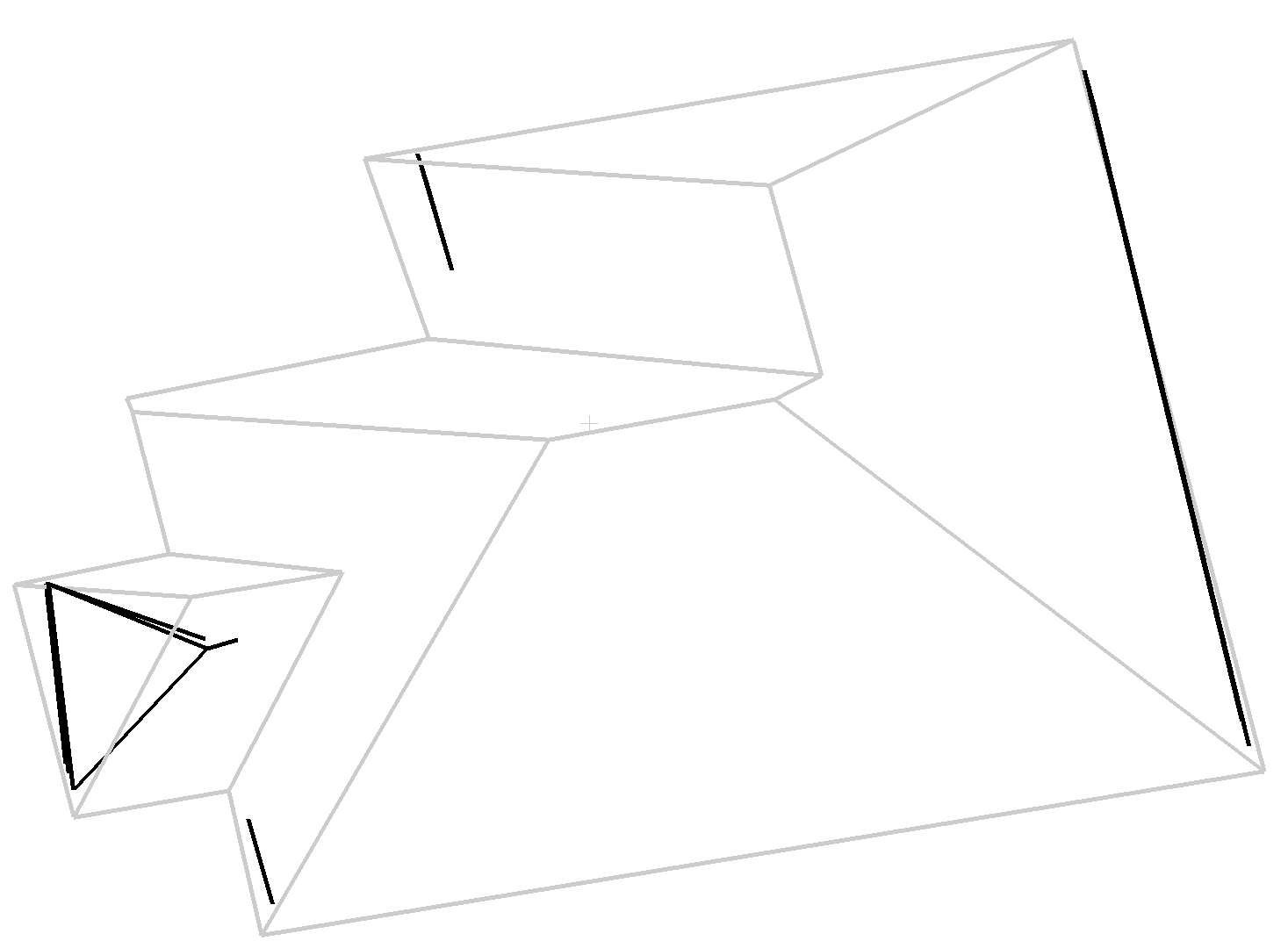}
        \includegraphics[width=\linewidth]{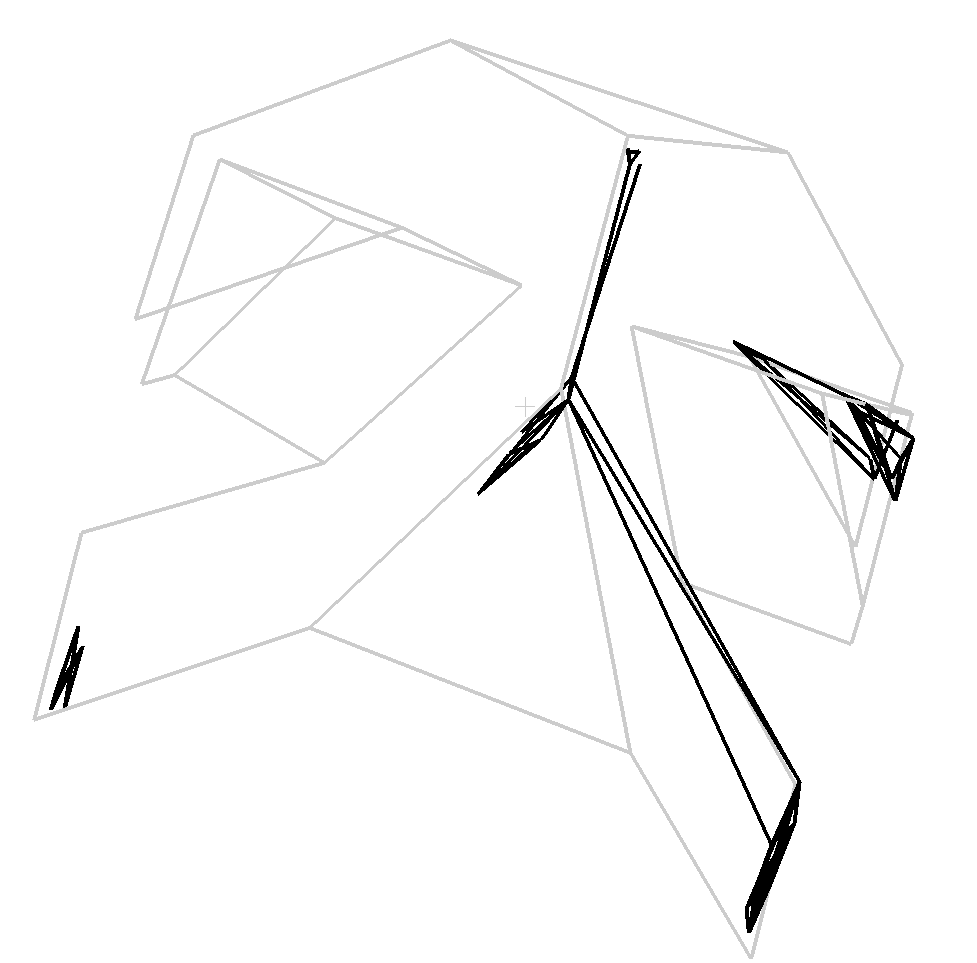}
        \includegraphics[width=1\linewidth]{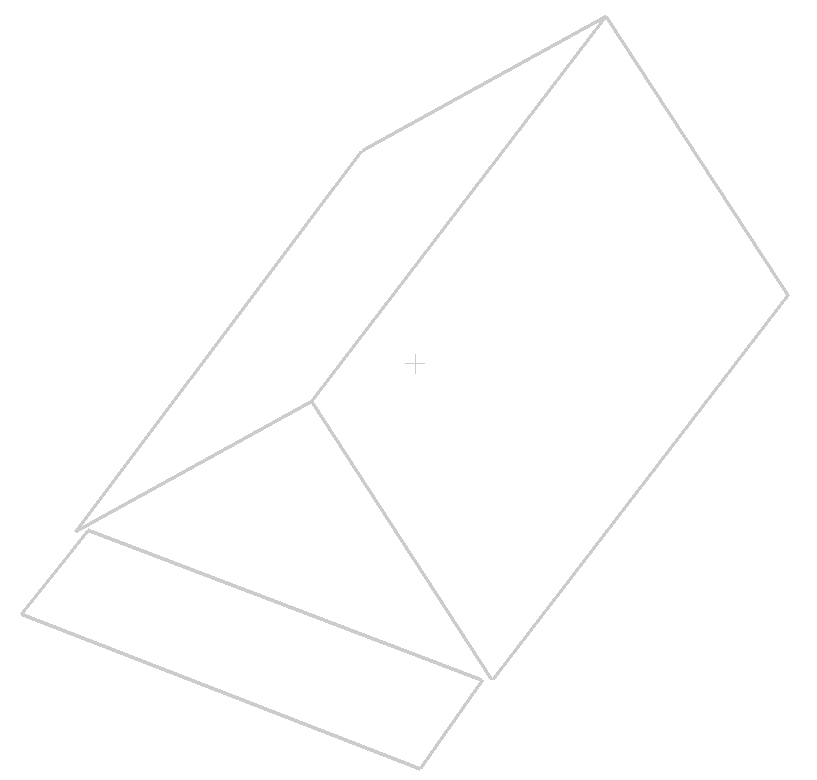}
        \includegraphics[width=\linewidth]{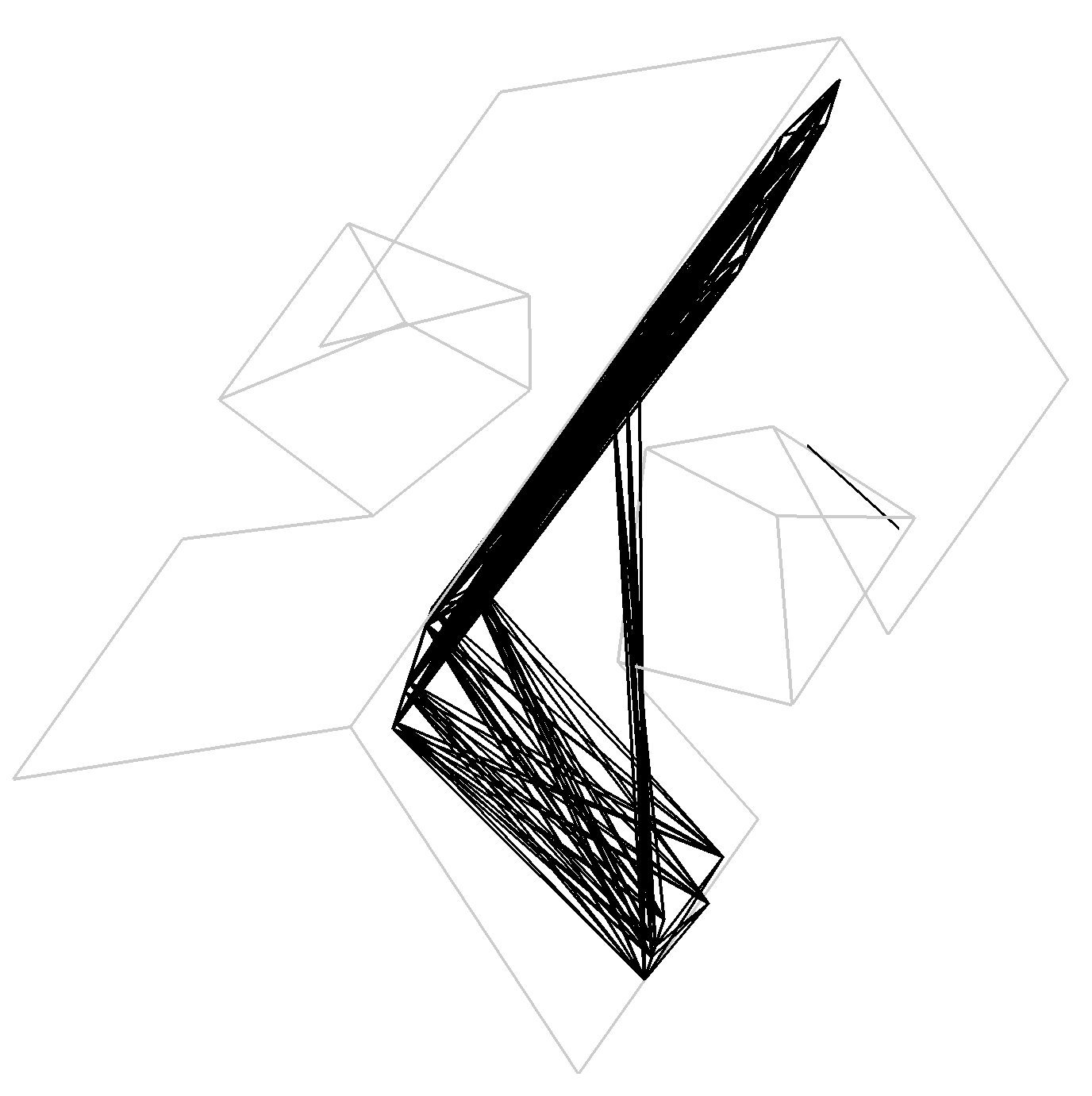}
        \includegraphics[width=1.1\linewidth]{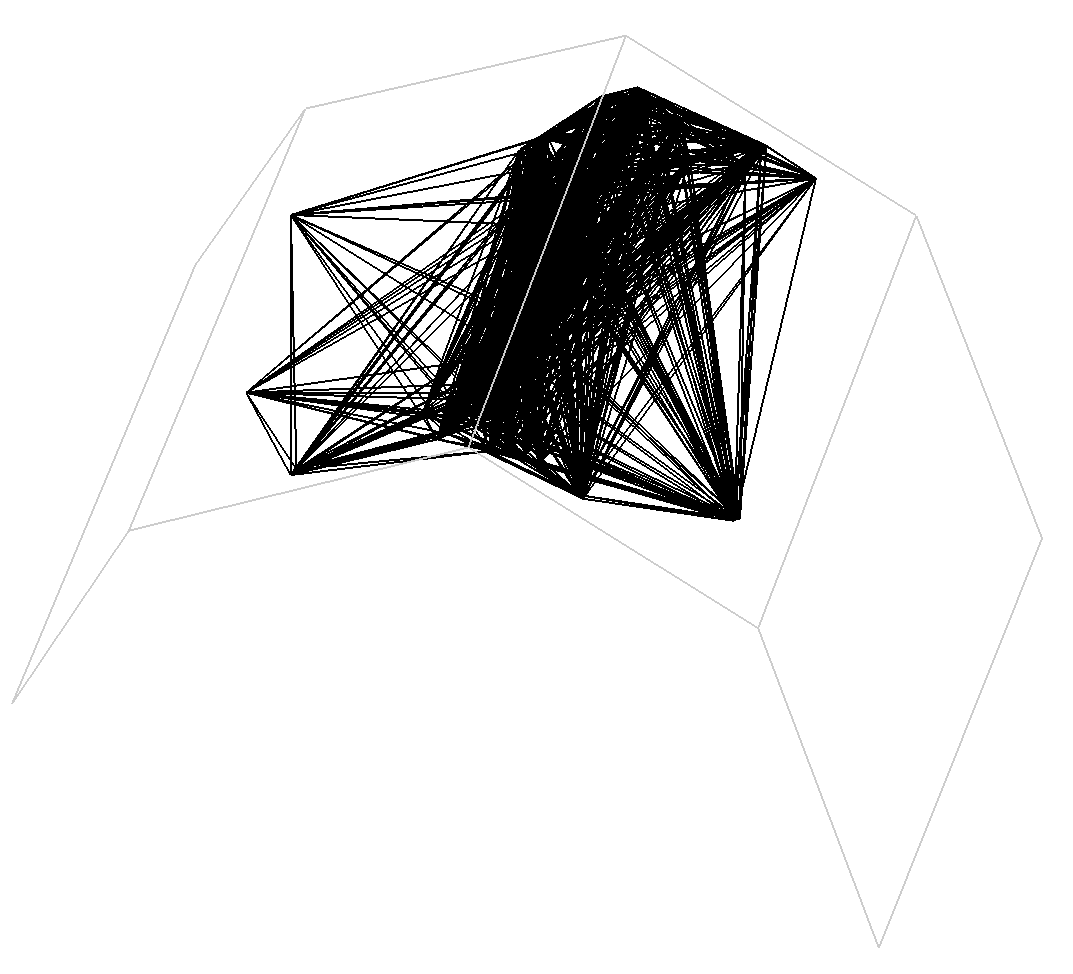}
        \includegraphics[width=\linewidth]{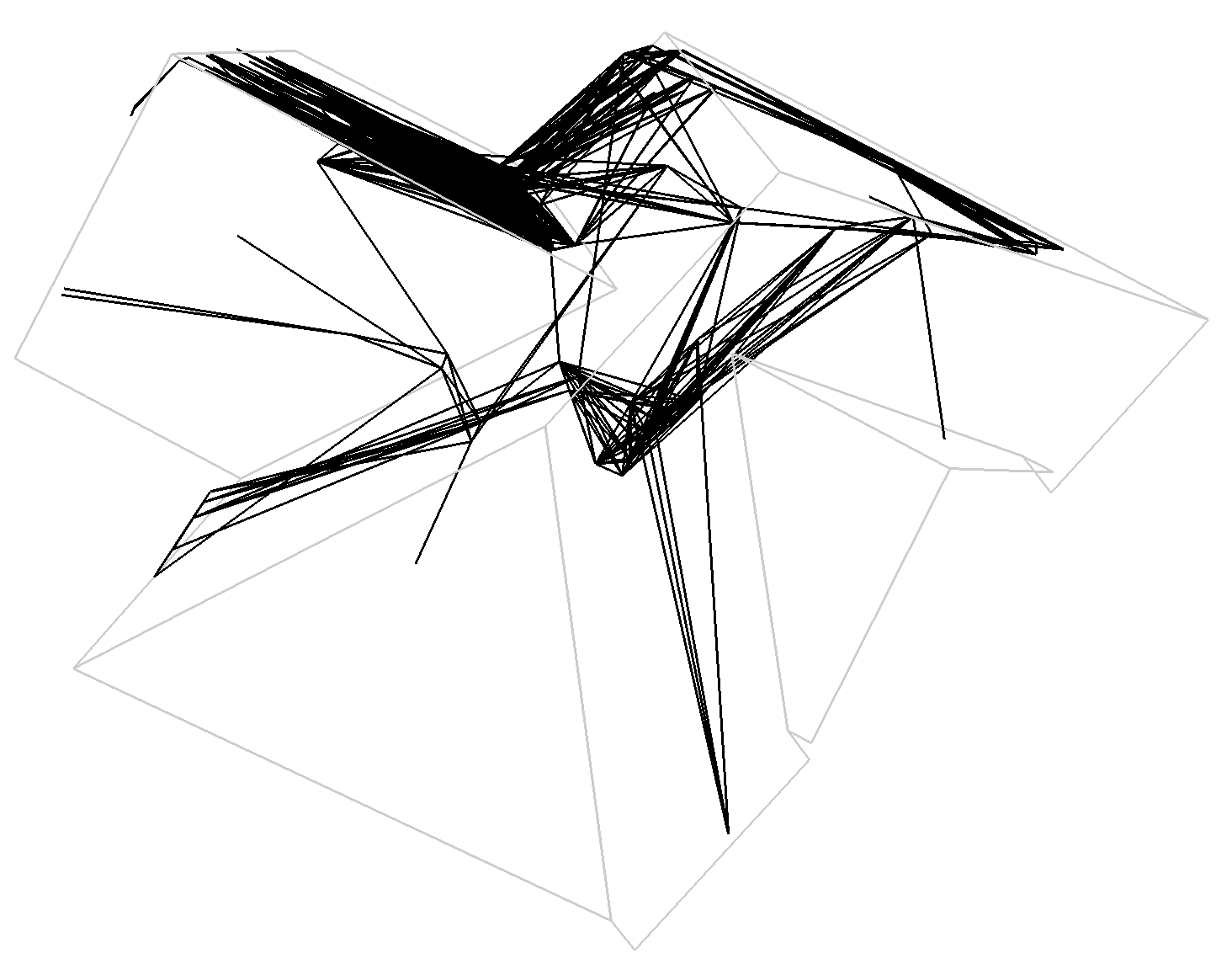}
        \includegraphics[width=\linewidth]{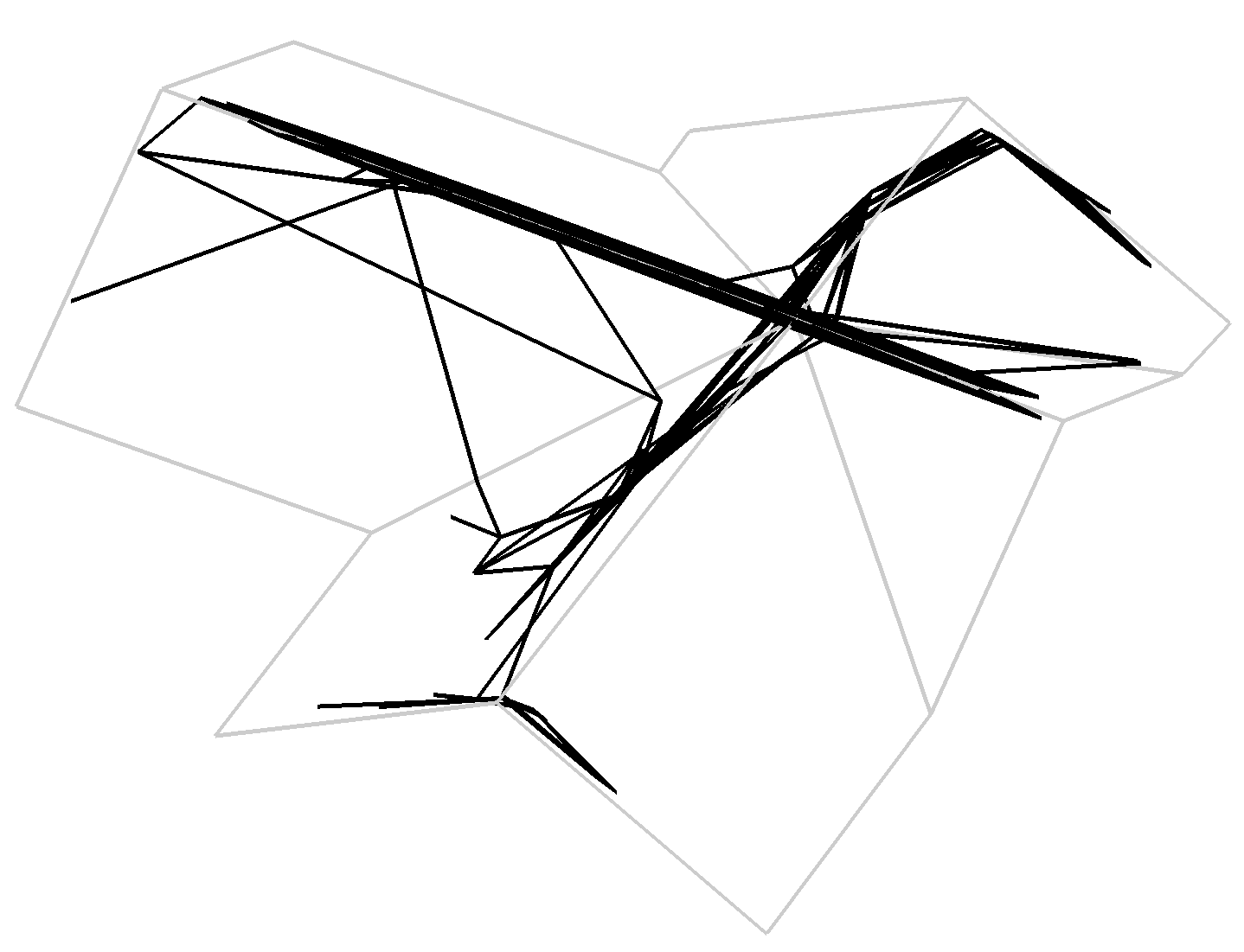}
        \includegraphics[width=\linewidth]{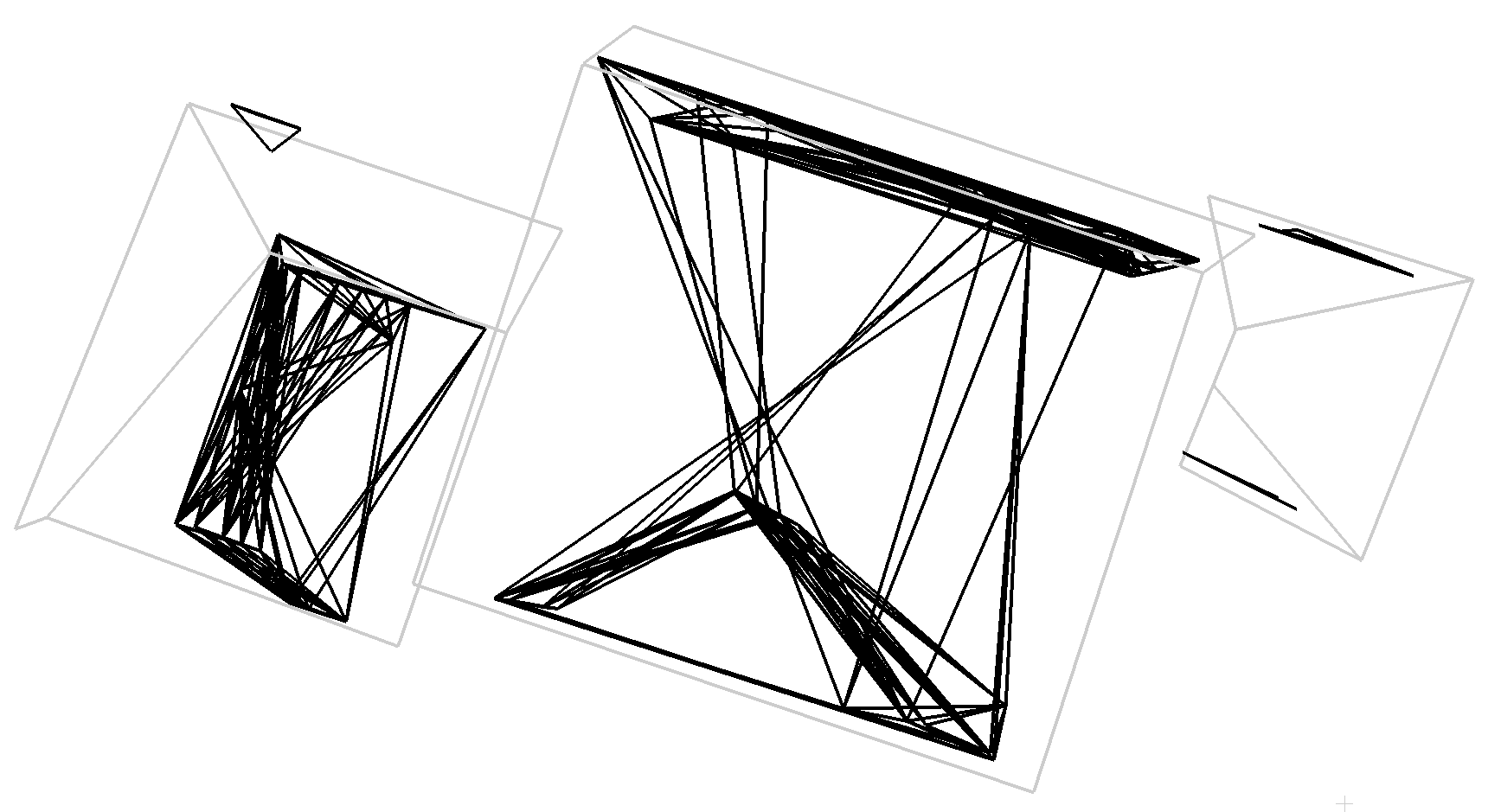}
        \subcaption*{\parbox{\linewidth}{\centering PC2WF\cite{liu2021pc2wf}}}
    \end{minipage}
    \hfill
    \begin{minipage}[t]{0.08\linewidth}
    \vspace{0pt}
        \includegraphics[width=\linewidth]{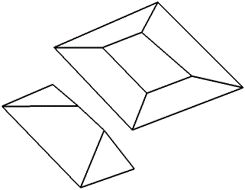}
        \includegraphics[width=\linewidth]{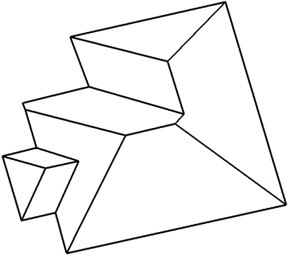}
        \includegraphics[width=\linewidth]{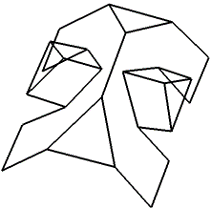}
        \includegraphics[width=0.9\linewidth]{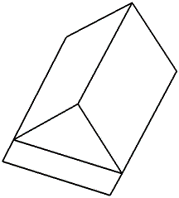}
        \includegraphics[width=\linewidth]{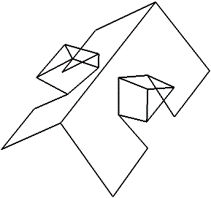}
        \includegraphics[width=0.92\linewidth]{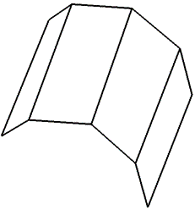}
        \includegraphics[width=\linewidth]{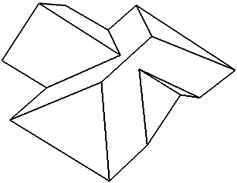}
        \includegraphics[width=\linewidth]{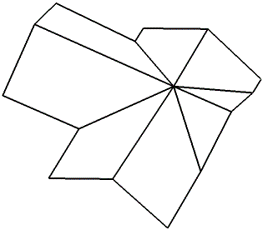}
        \includegraphics[width=\linewidth]{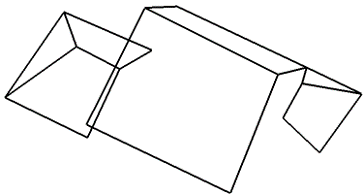}
        \subcaption*{\parbox{\linewidth}{\centering PBWR}}
    \end{minipage}
    \hfill
    \begin{minipage}[t]{0.08\linewidth}
    \vspace{0pt}
        \includegraphics[width=0.95\linewidth]{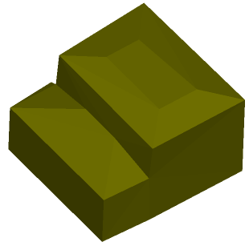}
        \includegraphics[width=0.9\linewidth]{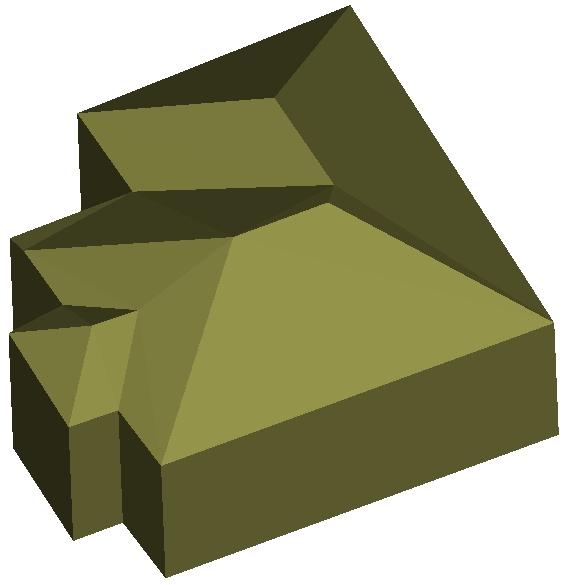}
        \includegraphics[width=\linewidth]{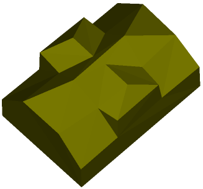}
        \includegraphics[width=0.9\linewidth]{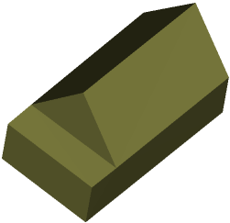}
        \includegraphics[width=\linewidth]{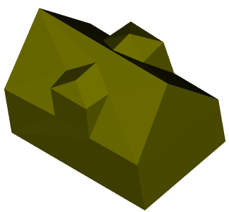}
        \includegraphics[width=0.87\linewidth]{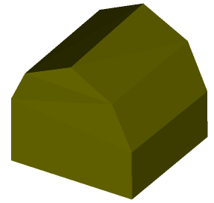}
        \includegraphics[width=0.9\linewidth]{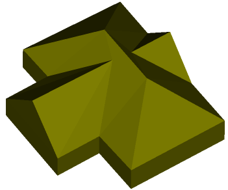}
        \includegraphics[width=0.9\linewidth]{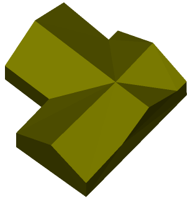}
        \includegraphics[width=0.9\linewidth]{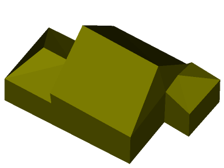}
        \subcaption*{\parbox{\linewidth}{\centering PBWR}}
    \end{minipage}
    \caption{Qualitative evaluation of traditional and deep learning methods. Traditional methods provide mesh results, and deep learning methods provide wireframe results. To enhance visualization clarity, unboldened ground truth data was incorporated as a background for better observation of the visualization results from EC-Net to PC2WF.}
    \label{fig_visual_results}
\end{figure*}
\begin{figure*}
    \centering
    \includegraphics[width=1\linewidth]{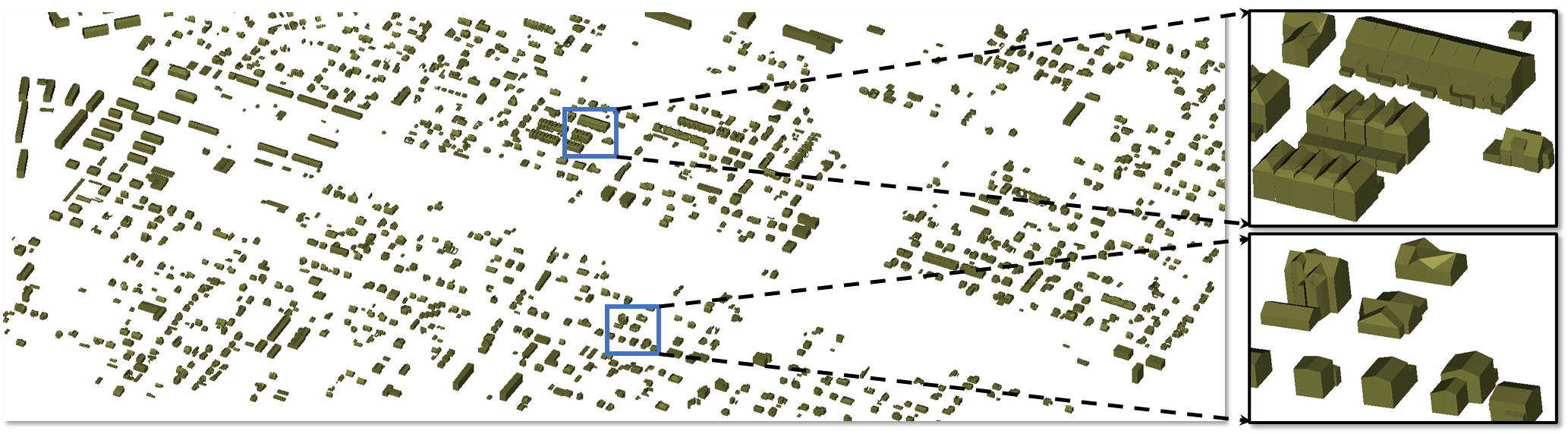}
    \caption{\textbf{Visualization of Tallinn City Data Reconstruction Results.} The wireframe reconstruction results of buildings in a 1000x2000 m area in central Tallinn city transformed into roof mesh and corresponding facade mesh}
    \label{fig:large}
\end{figure*}

\subsection{Implementation }
\label{Implementation}
The number of points $N$ within a batch is set to 2560, which is approximately equal to the average number of points per sample in the dataset.  The point features  $P \in \mathbb{R}^{N\times7}$ are fed into the input embedding module to obtain embedding features with $C_{embed} = 256$ channels. The output of encoder is features with $C_{en} = 256$ channels; then these features, along with $M=128$ query points and $C_{query} = 64$ query embedding are both fed into the decoder to obtain edge features $F_{edge}$ with $C_{edge} = 256$ channels. 128 query points is a reasonable setting as demonstrated in the ablation study.
Detailed parameter settings for the network, E-NMS, and other modules are provided in the \textit{supplementary materials}.

\subsection{Results and Comparisons}
\label{Results and Comparisons}
Through our best efforts, a summary of quantitative comparison between PBWR and existing wireframe reconstruction methods is presented in \cref{tab_entry_level} and \cref{tab_Tallinn}. Specifically, \cref{tab_entry_level} shows results on the Entry-level data, and \cref{tab_Tallinn} represents results on the Tallinn City data. * denotes that the method is employed as a feature extractor to extract point features, followed by the corner prediction and edge classification modules to reconstruct wireframe models as shown in \cref{fig:Pipeline Comparison}. Experimental results demonstrate that proposed PBWR exhibits a significant margin in performance compared to the baseline provided by Building3D, \textbf{CR and ER increase by 19\% and 36\%}, respectively, with corresponding $F_1$ scores showing improvement of 15\% and 25\% on Entry-level data. On the Tallinn City data, \textbf{CR and ER increase 15\% and 42\%}, respectively, with corresponding $F_1$ scores showing improvement of 14\% and 40\%. Furthermore, the ACO distance has significantly decreased. These remarkable improvements  are a result of PBWR's avoidance of the intermediate corner prediction step, leading to reduced error accumulation and fewer constraints on the edge regression performance. 

\begin{table}[h!]
    \centering
    \begin{tabular}{c|c|cc|cc}
    \toprule
          Method&  Dis.&\multicolumn{4}{c}{Accuracy}\\
  & ACO & CP& CR
& EP&ER
\\
 \midrule
  Corner Matching& 0.24& 0.97& 0.39& 0.93&0.30\\
 Midpoint Matching& 0.25& 0.96& 0.85& 0.87&0.81\\
 PBWR-Corner& 0.44& 0.90& 0.10& 0.06&0.12\\
    \rowcolor{blue!8}PBWR&  \textbf{0.22}&  \textbf{0.97}&  \textbf{0.85} &  0.91& \textbf{0.82} \\
    \bottomrule
    \end{tabular}
    \caption{Effect of difference bipartite edge matching strategies and regression methods}
    \label{tab_matching}
\end{table}

Some methods have been proposed recently for extracting edge points, which can be further processed for generating wireframe models. We present visual comparison with respect to these methods \cite{yu2018ec, zhu2023nerve, wang2020pie, liu2021pc2wf} and some popular traditional methods as shown in \cref{fig_visual_results}. We display extracted edges in black, some extracted corners in red, and ground truth wireframe model in gray to improve the visualization, especially when some compared methods can only extract incomplete edges.
The primary reason that these methods \cite{yu2018ec, zhu2023nerve, wang2020pie, liu2021pc2wf} fail is that their designed models do not adapt well to sparsity of aerial LiDAR point clouds, as well as the presence of missing data and noise in the point clouds. The quantitative comparison including RMSE, 3D IOU, and face number between PBWR and traditional methods are detailed in the \textit{supplementary  materials}. \cref{fig:large} visualizes results of a large-scale scene reconstruction.

\section{Ablation Study}
\label{sec:abla}
\begin{figure}[!ht]
    \centering
    \includegraphics[width=0.95\linewidth]{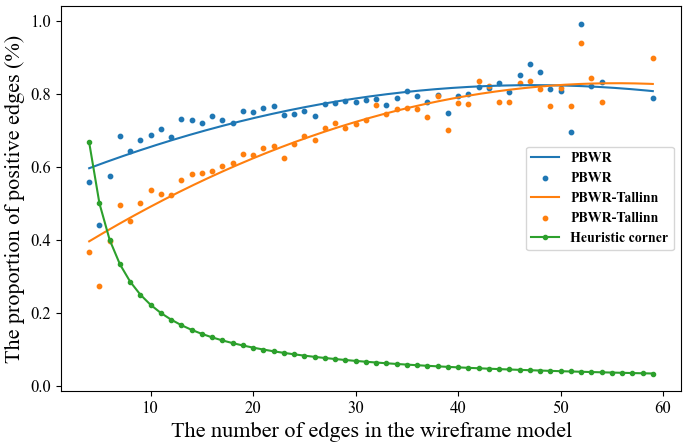}
    \caption{\textbf{Ratio of positive sample edges} Points represent the original data. The curve is fitted given the point set. Heuristic corner represents the popular method based on intermediate corner prediction.}
    \label{fig:positive}
\end{figure}
\noindent\textbf{Edge Generation} 
It is not advisable to use the strategy of directly regressing two endpoints of an edge (PBWR-Corner) as it results in poor performance as described in \cref{tab_matching}. Compared to these methods based on heuristic corner strategy, PBWR predicts more positive edges as shown in \cref{fig:positive}. As the number of these edges in a wireframe model increase, PBWR shows a rising ratio of predicted positive edges. However, the methods based on heuristic corner strategy, such as PC2WF\cite{liu2021pc2wf} and PointRoof \cite{li2022point2roof}, represents a continual declining trend. This significantly contributes to substantial improvement of performance demonstrated by our method.

\begin{figure}[!ht]
    \centering
    \includegraphics[width=0.48\linewidth]{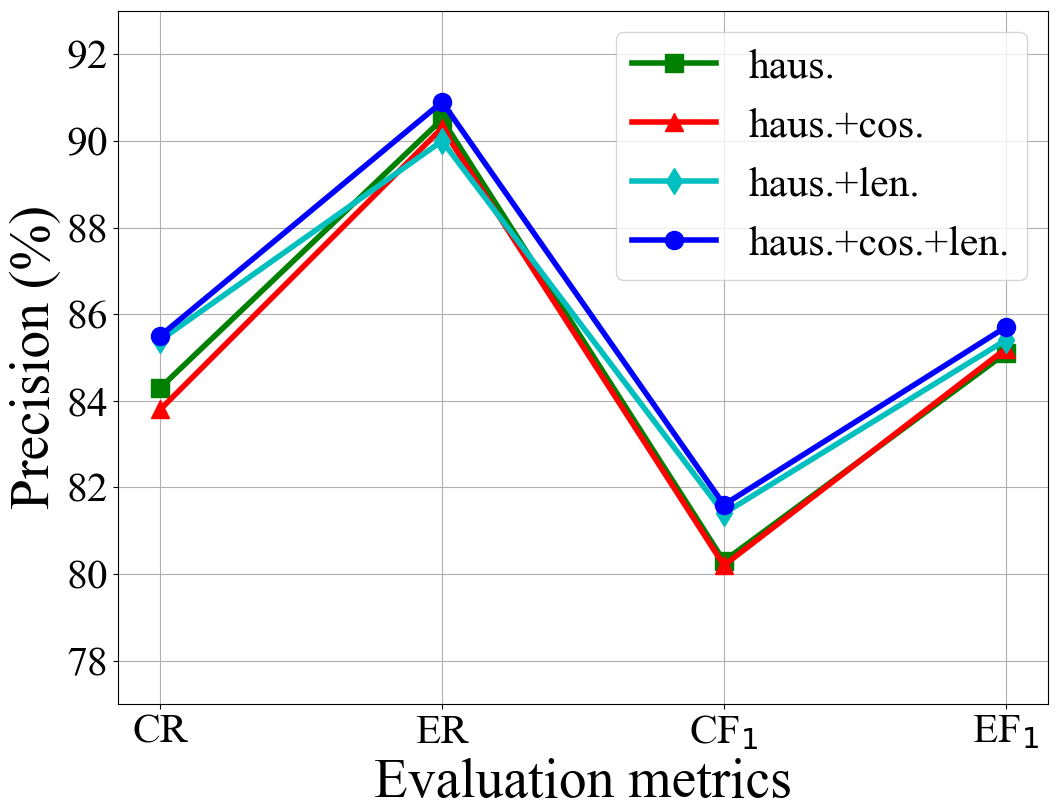}
    \includegraphics[width=0.48\linewidth]{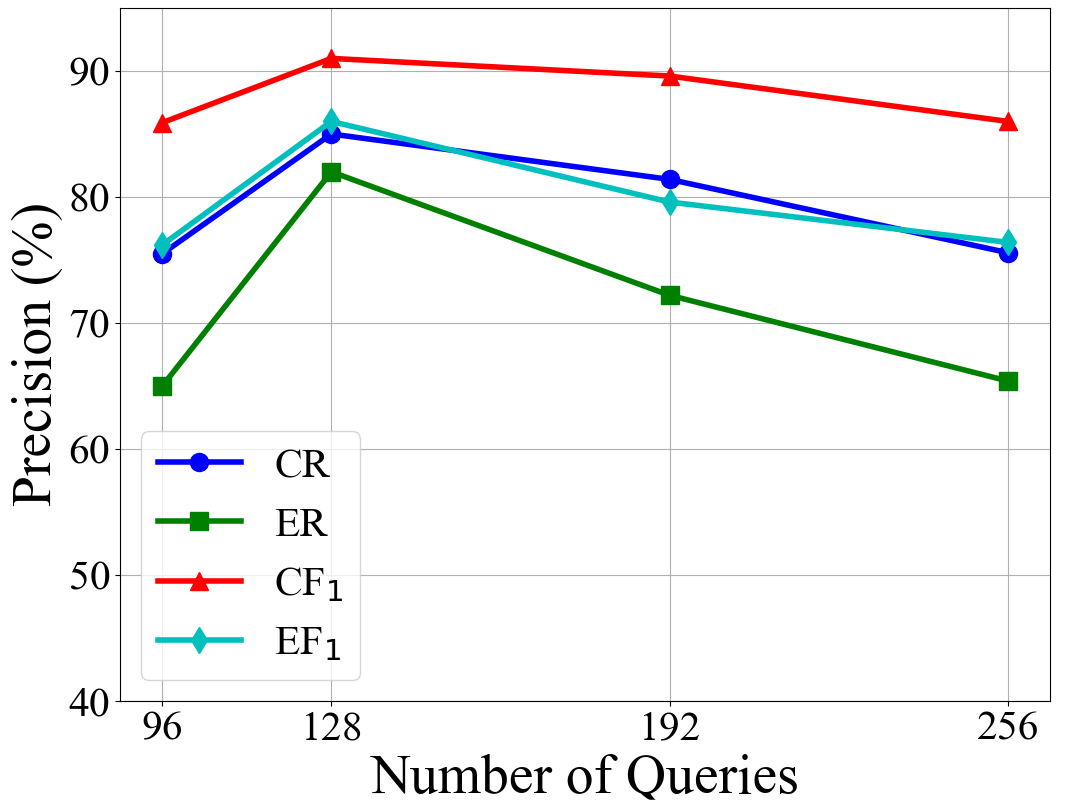}
    \caption{\textbf{Left}: Comparison of different combinations of matching strategies. \textbf{Right}: comparison of impact of four different query points on performance.}
    \label{fig:number_queries}
\end{figure}

\noindent\textbf{Bipartite Edge Matching} 
The experimental results in \cref{tab_matching} demonstrate that matching strategy by minimizing Hausdorff distances significantly outperforms strategies by matching corresponding corners and midpoints. Furthermore, the subtle performance differences using combination of Hausdorff distance, length and angle similarities are also shown in the \cref{fig:number_queries}. 

\noindent\textbf{E-NMS}
The performance of E-NMS module is demonstrated in the \cref{tab_effect_module}. The experimental results indicate that an excessive number of redundant positive sample edges, if not removed, result in a rapid decline in precision (\textbf{-36\% on CP, and -22\% on EP}) and also affect the recall.

\begin{table}[h!]
    \centering
    \begin{adjustbox}{width=\linewidth}
    \begin{tabular}{cccc|cc|cc}
    \toprule
             Hard&Soft &$\mathcal{L}_{sim}$&E-NMS&\multicolumn{4}{c}{Accuracy }\\Labels&Labels &&& CP& CR
& EP&ER
\\
 \midrule
     \checkmark &  &\checkmark& & 0.62& 0.42& 0.56&0.35
\\
 \checkmark&   &\checkmark& \checkmark& 0.98& 0.50
& 0.78&0.41
\\
 & \checkmark& & \checkmark& 0.97& 0.84& 0.89&0.80\\
       \rowcolor{blue!8}&\checkmark &\checkmark&\checkmark&  \textbf{0.97}&  \textbf{0.85} &  \textbf{0.91}& \textbf{0.82} \\
    \bottomrule
    \end{tabular}
    \end{adjustbox}
    \caption{\textbf{Effectiveness of modules}. Ablation study of different loss label strategies in the $\mathcal{L}_{con}$ loss and E-NMS module. }
    \label{tab_effect_module}
\end{table}

\noindent\textbf{Loss Module}
Previous bipartite edge matching results between prediction and ground truth can be utilized to assign ground truth labels to the prediction. Employing hard labels, commonly depicted as 0 and 1, is a prevalent practice. However, in doing so, the positive-to-negative sample ratio would be approximately 1:10.
In the confidence score loss, the extreme sample imbalance limits the performance of the network, even when the Focal loss is applied to hard labels as shown in \cref{tab_effect_module}. Hence, edge similarity is utilized as soft labels described in \cref{equ_conf}  to alleviate the sample imbalance.
In contrast to edge confidence scores, which necessitate clearly discriminative scores for positive and negative samples, other outputs only require optimization of obtained positive samples. Consequently, hard labels are employed to supervise these outputs. \cref{tab_effect_module} also demonstrates the significance of edge similarity loss $\mathcal{L}_{sim}$.

\noindent\textbf{Number of Queries}
\cref{fig:number_queries} depicts the influence of varying quantities of query points on performance. An insufficient number of points makes it impractical to generate enough predicted edges corresponding to the ground truth. Conversely, an excess of points results in an abundance of redundant edges, thereby increasing the network burden. 

\begin{figure}[!ht]
    \centering
    \includegraphics[width=0.18\linewidth]{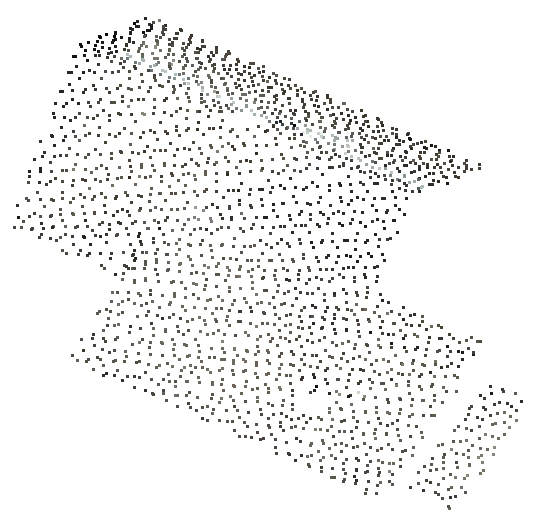}
    \includegraphics[width=0.19\linewidth]{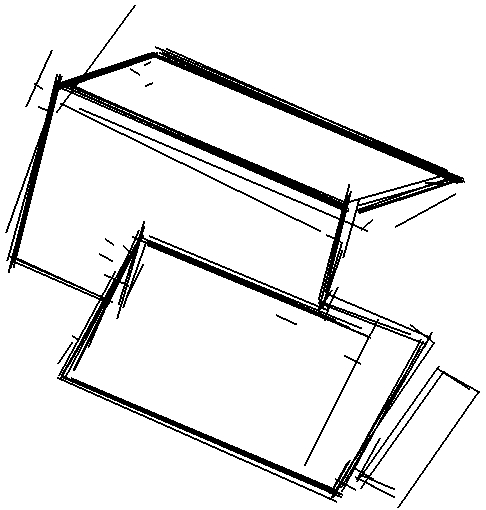}
    \includegraphics[width=0.19\linewidth]{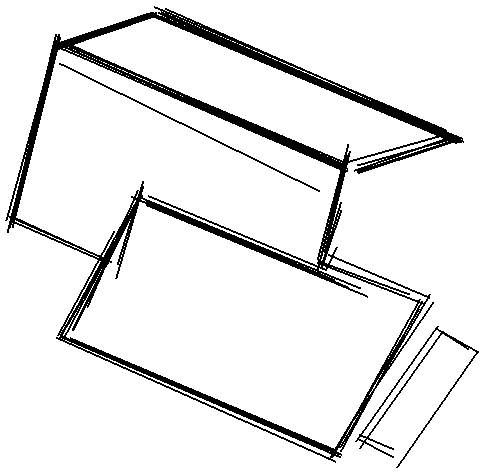}
    \includegraphics[width=0.19\linewidth]{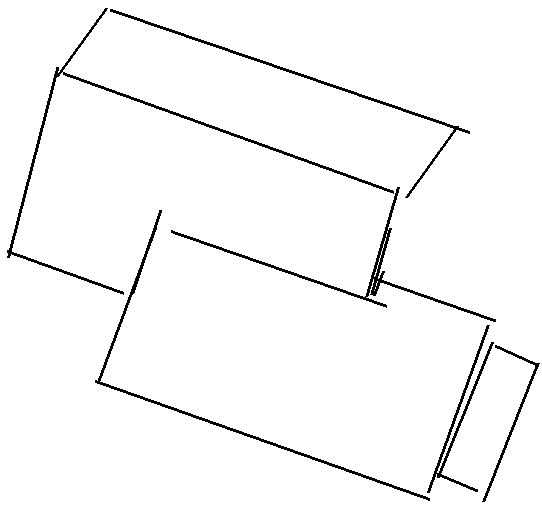}
    \includegraphics[width=0.19\linewidth]{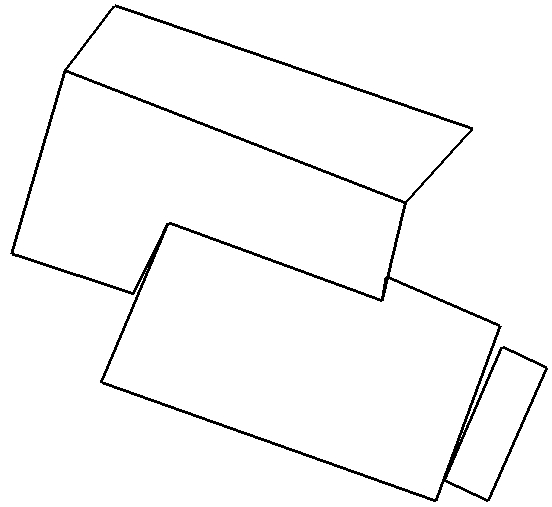}
    
    \caption{\textbf{Visualization of model generation.}: Left to right: point clouds, network output, confidence processing, E-NMS, and wireframe generation}
    \label{fig:model-generation}
\end{figure}

\noindent\textbf{Wireframe Model Generation}
During evaluation, the confidence score threshold is set to 0.7, effectively eliminating unexpected edges from the  original network output as shown in \cref{fig:model-generation}. The E-NMS algorithm is subsequently employed to eliminate remaining redundant edges. To obtain a seamlessly connected wireframe model, the DBSCAN algorithm \cite{ester1996density} is utilized to merge corners, using centroids of the clusters as new corners. All accuracy evaluation are based on the resultant wireframe models. Notes that the DBSCAN algorithm, with a distance threshold of 0.05 and a minimum point count of 2, is not used to generate any new edges.

\section{Conclusion}
\label{sec:conclusion}

In this paper, we propose PBWR, an end-to-end wireframe reconstruction model that directly regresses edges, bypassing intermediate heuristic modules for corner or edge prediction. The resultant parameterized edges undergo bipartite edge matching using the proposed Hausdorff distance-based similarity algorithm. E-NMS leverages edge similarity as a crucial parameter to eliminate redundant positive sample edges.
Additionally, a loss function specifically designed for edge optimization is proposed to guide the network optimization and model generation. In experiments, PBWR achieves performance far beyond existing baselines.

\noindent\textbf{Limitations and Future Work} One limitation is that the edges generated by PBWR are not a continuously connected edge set as shown in \cref{fig:model-generation}. Fortunately, corners of the obtained edge sets are closely located. Therefore, the DBSCAN algorithm \cite{ester1996density} with a small distance threshold of 0.05 m is employed to aggregate endpoints. In our opinion, this strategy might not be prudent, as it will result in unforeseen errors. In our upcoming research, our emphasis will be on eliminating the need for DBSCAN processing and instead directly generate continuously connected set of edges.



{\small
\bibliographystyle{ieee_fullname}
\bibliography{egbib}
}

\end{document}